\documentclass[10pt,twocolumn,letterpaper]{article}

\usepackage{cvpr}              
\usepackage[table,dvipsnames]{xcolor}

\usepackage{bbding}

\usepackage{enumitem}
\usepackage{tabularx}
\usepackage{colortbl}
\usepackage{booktabs}
\usepackage{multirow}
\usepackage{array}
\usepackage{graphicx}
\usepackage{caption}
\usepackage{adjustbox} 
\usepackage{makecell}
\usepackage[accsupp]{axessibility}
\usepackage[dvipsnames]{xcolor}

\definecolor{cvprblue}{rgb}{0.21,0.49,0.74}
\usepackage[pagebackref,breaklinks,colorlinks,citecolor=cvprblue]{hyperref}
\newcommand{\ours}{ULDA}

\newcommand{\mypara}[1]{\vspace{0.05cm}\noindent\textbf{#1}\hspace{0.1cm}}
\renewcommand{\vec}[1]{\boldsymbol{#1}}
\newcommand{\DAsetting}[2]{{#1}$\rightarrow${#2}}
\def\paperID{7095} 
\def\confName{CVPR}
\def\confYear{2024}

\title{Unified Language-driven Zero-shot Domain Adaptation}

\author{Senqiao Yang\textsuperscript{1,2}  \quad Zhuotao Tian\textsuperscript{2}\thanks{Corresponding Author (\href{tianzhuotao@hit.edu.cn}{ tianzhuotao@hit.edu.cn}).} \quad Li Jiang\textsuperscript{3}\quad Jiaya Jia\textsuperscript{1} \\
\textsuperscript{1}The Chinese University of Hong Kong\quad \\
\textsuperscript{2}Harbin Institute of Technology, Shenzhen\quad \textsuperscript{3}The Chinese University of Hong Kong, Shenzhen 
}

\begin{document}

\maketitle
\begin{abstract}
This paper introduces Unified Language-driven Zero-shot Domain Adaptation (ULDA), a novel task setting that enables a single model to adapt to diverse target domains without explicit domain-ID knowledge. We identify the constraints in the existing language-driven zero-shot domain adaptation task, particularly the requirement for domain IDs and domain-specific models, which may restrict flexibility and scalability. To overcome these issues, we propose a new framework for ULDA, consisting of Hierarchical Context Alignment (HCA), Domain Consistent Representation Learning (DCRL), and Text-Driven Rectifier (TDR). These components work synergistically to align simulated features with target text across multiple visual levels, retain semantic correlations between different regional representations, and rectify biases between simulated and real target visual features, respectively. 
Our extensive empirical evaluations demonstrate that this framework achieves competitive performance in both settings, surpassing even the model that requires domain-ID, showcasing its superiority and generalization ability. The proposed method is not only effective but also maintains practicality and efficiency, as it does not introduce additional computational costs during inference. The code is available on the project website\footnote{\href{https://senqiaoyang.com/project/ULDA}{Senqiaoyang.com/project/ULDA}}.

\begin{figure}[!t]
    \centering
    \includegraphics[width=.9\linewidth]{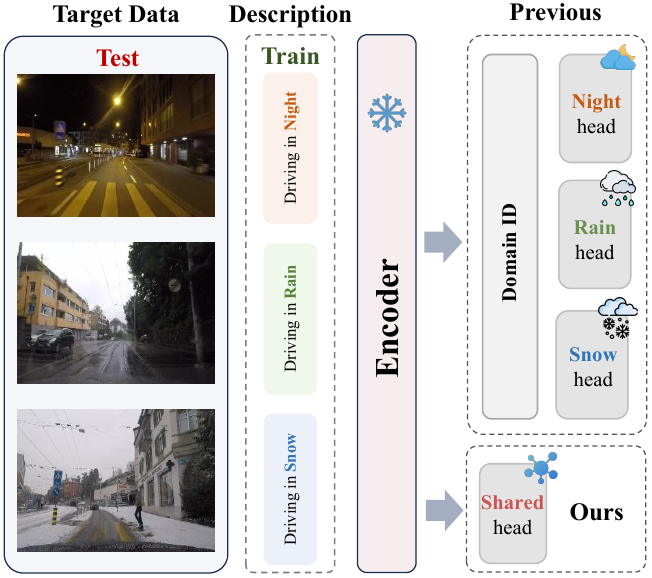}
    \caption{Our proposed Unified Language-driven Domain Adaptation (ULDA) task focuses on real-world practical scenarios. In the training phase, ULDA does not allow access to images of the target domain and only provides source domain images along with the textual descriptions.
    During testing, ULDA requires a single model to adapt to diverse target domains without domain-IDs, instead of using domain-specific heads as in previous methods.
    }
    \label{fig:intro}
    \vspace{-0.45cm}
\end{figure}

\end{abstract}
\section{Introduction}
Being robust to the domain shift is a critical concept in machine learning, as it enables models trained on a source domain to be effectively applied to a new target domain~\cite{ tsai2018learning, hoffman2018cycada, ge2022domain, yang2020fda}.
The domain adaptation (DA) task~\cite{decouplenet, lin2022prototype, hoyer2022hrda, ding2022source, wang2022exploring} may assume the availability of target domain data for fine-tuning the model. However, this assumption may potentially hinder real-world applications~\cite{liu2023vida, yang2024exploring, benigmim2023one, fahes2023poda}. For example, privacy concerns or data scarcity may prevent direct access to target data. Therefore, the underlying challenges caused by the absence of direct access to target domain data in developing the domain-versatile models should be considered and decently addressed, ensuring the applicability of DA techniques in practical situations.

Recently, the development of vision-language foundational models~\cite{radford2021learning, ramesh2021zero, alayrac2022flamingo, lai2023lisa, yang2023improved} has greatly advanced the alignment of image-text pairs, enabling effective generalization to novel concepts. This has paved the way for numerous studies that leverage the zero-shot capabilities of these models to tackle domain adaptation challenges~\cite{radford2021learning, kwon2021clipstyler,gal2022stylegan,kwon2023one, ge2022domain}. Notably, P{\O}DA\xspace~\cite{fahes2023poda} stands out for leveraging language embeddings obtained from CLIP~\cite{radford2021learning} to simulate target domain visual representations, and P{\O}DA\xspace tunes the model to fit the simulated features, such that the target images are not needed. 

However, while P{\O}DA\xspace demonstrates an impressive ability to achieve zero-shot domain adaptation without relying on target domain images, we observe that it introduces constraints that should be taken into account in practical contexts: \textit{prior knowledge, \textit{i.e.}, domain-ID, is required to select the domain-specific model}. For example, for ``driving in rain'' and ``driving in snow'', two individual models are trained to fit these two task domains with the help of CLIP text embeddings. During inference, when an image comes, the system processes in two steps: 1) know what the current domain is via Domain ID, and then 2) select the corresponding model.
These domain-specific customization requirements may hinder the model's flexibility and scalability, thereby limiting its broader applications.

To address the aforementioned issue, we propose a novel and practical task called Unified Language-driven Zero-shot Domain Adaptation (ULDA), as shown in Fig.~\ref{fig:intro}. Following previous literature~\cite{fahes2023poda}, ULDA also does not allow access to images of the target domain, only providing source domain images along with the textual descriptions regarding target domains to accomplish the adaptation training process. 
However, ULDA takes a step further by requiring a single model to adapt to diverse target domains without explicit hints, \textit{i.e.}, the domain-ID, during testing. Nevertheless, this new requirement also presents a significant challenge: \textit{how to adapt a single model's embedding space to accommodate multiple domains while still maintaining semantic discriminative capabilities for different categories?}

To address this challenge, we propose a new framework for ULDA. It has three essential components: Hierarchical Context Alignment (HCA), Domain Consistent Representation Learning (DCRL), and Text-Driven Rectifier (TDR). Specifically, HCA aligns simulated features with target text at multiple visual levels to mitigate the semantic loss caused by the vanilla scene-text alignment. 
Then, DCRL retains the semantic correlations between different regional representations to that of the text embeddings across diverse domains, ensuring structural consistency.
Additionally, we incorporate TDR to rectify simulated features, mitigating the bias between the simulated and real target visual features.

We validate the effectiveness of our proposed method through extensive empirical evaluations conducted in both the previous classic setting~\cite{fahes2023poda} and the proposed ULDA. The results consistently demonstrate that our approach achieves competitive performance in both settings, highlighting its superiority and efficacy. In summary, our contribution can be summarized in three key aspects:
\begin{itemize}
    \item Unlike existing literature, we go beyond existing approaches by examining the limitations that hinder further applications. To this end, we propose a more practical setting called Unified Language-driven Zero-shot Domain Adaptation (ULDA).
    \item To address the new challenge posed by ULDA, we propose a new framework, and it comprises three key components, namely Hierarchical Context Alignment (HCA), Domain Consistent Representation Learning (DCRL), and Text-Driven Rectifier (TDR), for achieving better alignment to the text embedding space, ensuring a better adaptation performance.
    \item Despite its simplicity, our proposed method's effectiveness has been verified in both settings. Furthermore, it does not introduce any additional computational costs during model inference, ensuring its practicality. 
\end{itemize}

\section{Preliminary}
\label{sec:preliminary}
In this section, we introduce a closely related work P{\O}DA\xspace~\cite{fahes2023poda}, which proposes a paradigm for prompt-driven zero-shot domain adaptation in computer vision, by only leveraging a natural language description of the target domain, thus eliminating the need for target domain images during training. A more detailed introduction regarding related works is shown in the supplementary file.

Specifically, P{\O}DA\xspace undergoes two stages of training to leverage the pretrained CLIP encoder for optimizing source feature transformations and aligning them with the text embedding of the target domain. In Stage-1, it learns to simulate target features. In Stage-2, it fine-tunes the segmentation head with the simulated ones. Details are as follows.

\textbf{Stage-1:} P{\O}DA\xspace introduces Prompt-driven Instance Normalization (PIN), as in Eq.~\eqref{eq:pin}, in which $\boldsymbol{\mu}$ and $\boldsymbol{\sigma}$ are learnable variables guided by a text domain prompt to simulate the knowledge of the target domain, while $\mu(\mathbf{f}_{\mathrm{s}})$ and $\sigma(\mathbf{f}_{\mathrm{s}})$ represent the mean and standard deviation of the source input features~$\mathbf{f}_{\mathrm{s}}$.

\begin{equation}
\label{eq:pin}
\mathbf{f}_{\mathrm{s} \rightarrow \mathrm{t}} = \operatorname{PIN}\left(\mathbf{f}_{\mathrm{s}}, \boldsymbol{\mu}, \boldsymbol{\sigma}\right)=\boldsymbol{\sigma}\left(\frac{\mathbf{f}_{\mathrm{s}}-\mu\left(\mathbf{f}_{\mathrm{s}}\right)}{\sigma\left(\mathbf{f}_{\mathrm{s}}\right)}\right)+\boldsymbol{\mu}.
\end{equation}
PIN is adopted to transform source domain features into the target domain, \textit{i.e.}, $\mathbf{f}_{\mathrm{s} \rightarrow \mathrm{t}}$. This operation is followed by an attention-based pooling operation, resulting in the output denoted as $\overline{\mathbf{f}}_{\mathrm{s} \rightarrow \mathrm{t}}$. To ensure a proper shift from the source to the target domain, it is necessary to promote similarity between $\overline{\mathbf{f}}_{\mathrm{s} \rightarrow \mathrm{t}}$ and CLIP text embeddings $\operatorname{TrgEmb}$, by applying the loss function presented in Equation~\eqref{eq:similarity}, which encourages alignment between the transformed features and the textual representations, facilitating decent adaptation from the source to target domain.

\begin{equation}
\label{eq:similarity}
\mathcal{L}_{\mu, \sigma}\left(\overline{\mathbf{f}}_{\mathrm{s} \rightarrow \mathrm{t}}, \operatorname{TrgEmb}\right)=1-\frac{\overline{\mathbf{f}}_{\mathrm{s} \rightarrow \mathrm{t}} \cdot \operatorname{TrgEmb}}{\left\|\overline{\mathbf{f}}_{\mathrm{s} \rightarrow \mathrm{t}}\right\|\|\operatorname{TrgEmb}\|} .
\end{equation}

\textbf{Stage-2:}  With the simulated features obtained via Eq.~\eqref{eq:pin}, P{\O}DA\xspace fine-tunes the pre-trained segmentation head to enable the model to better adapt to the target domain for accomplishing the downstream task. This stage's training is supervised by the cross-entropy loss between the segmentation predictions and the ground-truth masks. 

It is worth noting that, for both two training stages, only the images from the source domain and text descriptions are available. After two phases of training, the model is evaluated on the images of target domains. More details can be found in~\cite{fahes2023poda}.

\section{ULDA: Unified Language-driven Zero-shot Domain Adaptation}
\label{Sec: task settings}

\textbf{Motivation. \quad} Traditional domain adaptation methods often depend on having access to data from the target domain in order to align the models. However, this dependence on target domain data can lead to overfitting to specific domains and subsequently undermine the generalization performance of the models. In real-world applications, especially in dynamic settings like autonomous driving, acquiring comprehensive data for every possible adverse condition (e.g., rain, snow, fog, night) is not always feasible. Instead, practitioners may only have a conceptual understanding or hypothetical descriptions of potential downstream tasks. 
In this case, the ability to augment a model's performance in such predicted scenarios without actual data collection is preferred. 

To achieve this, P{\O}DA\xspace~\cite{fahes2023poda} tunes different models to tackle individual scenarios separately. However, the prior knowledge of each upcoming domain for selecting the corresponding model may not always be accessible in practice. Therefore, we believe it is necessary to adopt an adaptation approach such that a single model can be scaled to simultaneously fit a broad spectrum of domain conditions. 

Considering these practical constraints, we propose a new task setting, namely Unified Language-driven Zero-shot Domain Adaptation (ULDA), which encourages the model to be capable of adapting to a variety of conditions without access to real data and domain prior knowledge.

\mypara{Task setting.}  
The model $\mathcal{M}$ is limited to training with data $\mathcal{I}_s$ from the source domain $\mathcal{D}_s$ and has no access to target domain data $\mathcal{I}_{t}$, where $t=1,2...n$ represent the $n$ target domains $\mathcal{D}_{t}$.
$\mathcal{M}$ can only utilize natural descriptions $\mathcal{T}_{t}$ to understand the characteristics of target domain scenarios. 

One of the challenges in this context is to effectively extract sufficient information from textual descriptions alone for adapting source visual features to different domains. Another crucial challenge is to enable a single model $\mathcal{M}$ to adapt to multiple target domains $\mathcal{D}_{t}$ without relying on specific domain IDs. This would allow the model to achieve robustness across diverse scenarios while still maintaining a strong ability to discriminate between different classes.

\mypara{Comparison with other settings.}
Different from Unsupervised Domain Adaptation (UDA), ULDA offers the advantage of generalizing to target domains without the need for target domain images. Instead, it only relies on a concise one-sentence description for each domain. This leads to a significant reduction in resource overhead. Additionally, unlike prompt-driven zero-shot domain adaptation proposed in P{\O}DA\xspace that requires domain IDs to invoke domain-specific models, the proposed ULDA enables a single model to adapt to multiple downstream scenarios without the requirement for separate tuning for each scenario.

\begin{figure*}
    \centering
    \includegraphics[width=.75\linewidth]{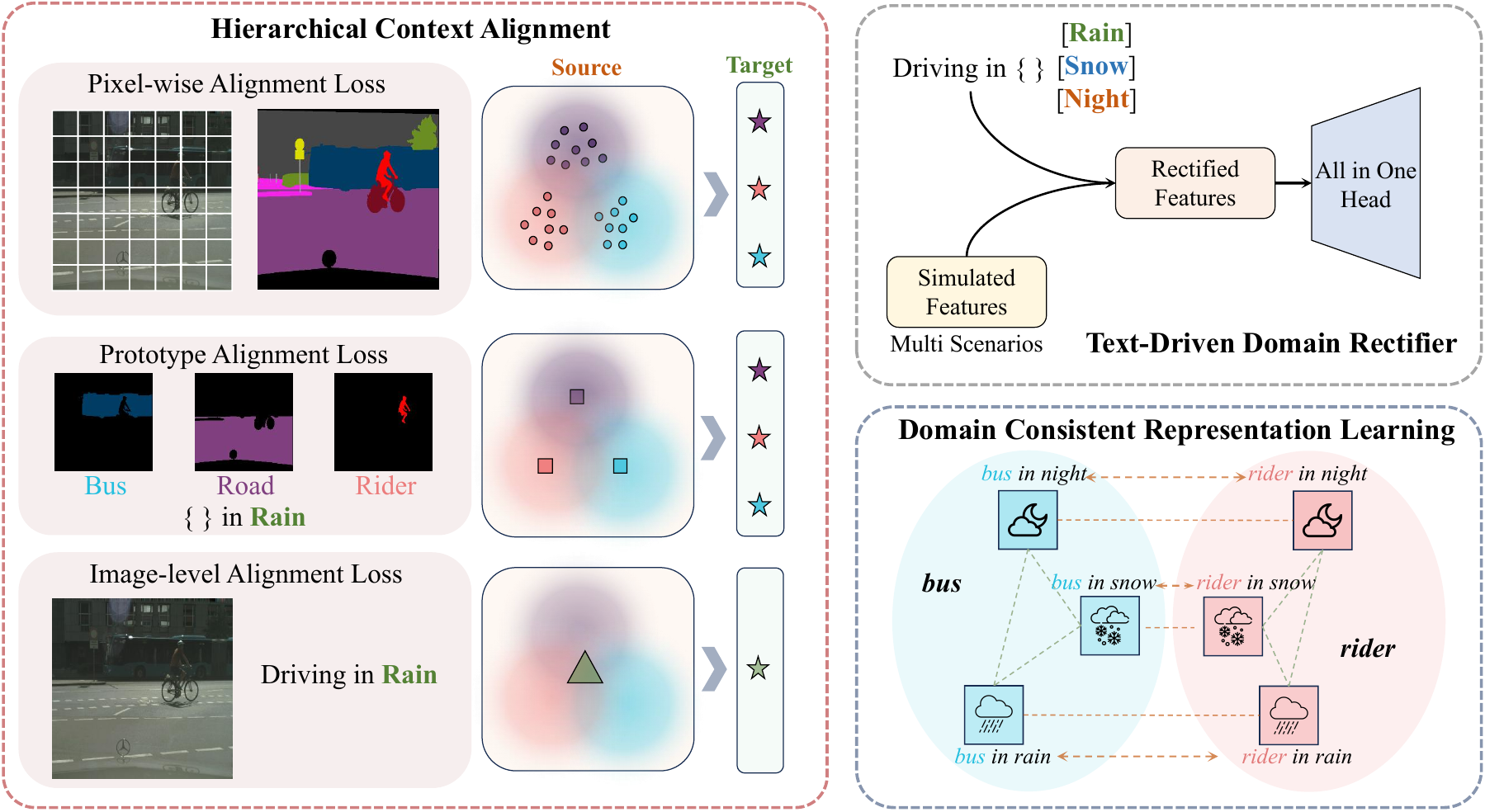}
    \caption{\textbf{Illustration of the three key components for our \ours~framework.} The ULDA's pipeline follows~\cite{fahes2023poda}. Our proposed Hierarchical Context Alignment operates across Pixel-level, Regional-level, and Scene-level to align features with text embeddings. The circles, squares, and triangles represent the hierarchical features, respectively. The Domain Consistent Representation Learning ensures a consistent correlation between prototypes and text embeddings across multiple target domains. Text-Driven Rectifier incorporates text embeddings to rectify the simulated PIN features during the fine-tuning phase.
    }
    \label{fig:ingro}
    \vspace{-0.5cm}
\end{figure*}

\section{Method}
The proposed ULDA brings a challenge in representation learning as a single model needs to adapt to multiple domains. This challenge arises from the fact that aligning the model towards target domains, such as ``driving in rain" and ``driving in snow," may potentially compromise semantic discrimination for precise segmentation.

For better accomplishing ULDA, we propose a framework that is composed of three components. The overview is shown in Fig.~\ref{fig:ingro}, and the respective details are as follows.

\subsection{Hierarchical Context Alignment}
\mypara{Vanilla scene-text alignment causes semantic loss.}
P{\O}DA\xspace achieved vision-language alignment at a scene level by directly aligning the pooled feature $\overline{\mathbf{f}}_{\mathrm{s} \rightarrow \mathrm{t}}$ with the text embedding $\operatorname{TrgEmb}$ via Eq.~\eqref{eq:similarity}. 
However, it is challenging for the model to achieve a decent alignment with the target domain by only adapting the global context to fit the target domain, because this may cause potential semantic loss when aligning features of different objects in a scene to a single shared target text domain embedding, causing a deviation from their respective real semantic distributions.

To alleviate this issue, we propose a Hierarchical Context Alignment (HCA) strategy, which enables intricate alignments on the feature $\mathbf{f}_{\mathrm{s} \rightarrow \mathrm{t}}$ at multiple levels, including 1) the entire scene, 2) regions in the scene, and 3) pixels in the scene.
The scene-text alignment follows that of Eq.~\eqref{eq:similarity}, while the proposed region- and pixel-text alignments are elaborated as follows.

\mypara{Regional alignment.} 
During the adaptation process, it is essential for regions belonging to different categories to retain their unique semantic characteristics. To achieve this, by leveraging the class names existing in the ground truth, and the target domain description, we can get the more fine-grained $d$-dimensional text embedding $\mathcal{T}\in \mathcal{R}^{[n \times d]}$ of $n$ classes contained in the image. By doing so, we can align different regions with more suitable counterparts, ensuring that their individual semantic characteristics are preserved.
For example, in a rainy scenario with $n$ classes, we can get descriptions such as ``the bus in rain," ``the road in rain," ``the rider in rain," and so on. Then, the corresponding text embeddings $\mathcal{T}\in \mathcal{R}^{[n \times d]}$ for these descriptions can be obtained from the pre-trained CLIP text encoder.

After that, given the image feature map $\mathbf{f}_{\mathrm{s} \rightarrow \mathrm{t}} \in \mathcal{R}^{[HW \times d]}$ and text embedding $\mathcal{T}\in \mathcal{R}^{[n \times d]}$, the pixel-wise ground-truth annotation $\vec{y} \in \mathcal{R}^{[HW]}$ can be accordingly transformed into $n$ binary masks $\vec{y}_{*} \in \mathcal{R}^{[n \times HW]}$ 
indicating the existence of $n$ classes in $\vec{y}$. 
Then, we can obtain the regional prototypes $\mathcal{C} \in \mathcal{R}^{[n \times d]}$ by applying masked average pooling (MAP) with $\vec{y}_{*}$ and $\mathbf{f}_{\mathrm{s} \rightarrow \mathrm{t}}$ as Eq.~\eqref{eq:get_proto}: 
\begin{equation}
    \label{eq:get_proto}
    \small
    \mathcal{C} = \frac{\vec{y}_{*} \times \mathbf{f}_{\mathrm{s} \rightarrow \mathrm{t}}}{\sum_{j=1}^{HW} \vec{y}_{*}(\cdot, j)}.
\end{equation}
Then, we can calculate the cosine similarity matrix $\mathcal{S} \in \mathcal{R}^{[n \times n]}$ between the  $\mathcal{T}$ and  $\mathcal{C}$ in Eq.~\eqref{eq:similarity_proto}.
\begin{equation}
\label{eq:similarity_proto}
\small
\mathcal{S} =  \frac{ \mathcal{C} \times \mathcal{T}^T}{\| \mathcal{C}\| \, \|\mathcal{T}\|^T}
\end{equation}
Therefore, we could use the categorical prototypes $\mathcal{C}_y$ and text embeddings  $\mathcal{T}$ to accomplish the regional alignment as:
\begin{equation}
\label{eq:region_contras}
\small
\mathcal{L}_r=-\sum_{i=1}^n \log \left(\frac{\exp \left(\mathcal{S}_{i i} / \tau\right)}{\sum_{k=1}^n \exp \left(\mathcal{S}_{i k} / \tau\right)}\right)
\end{equation}
where $\tau$ is the temperature parameter, and we empirically set it to 0.1. Eq.~\eqref{eq:region_contras} encourages the regional prototypes to be similar to the corresponding text embeddings in the target domain while pushing away negative pairs.

\mypara{Pixel-wise alignment.} Building upon the regional alignment, we further enhance the alignment between the source and the unseen target domains by incorporating a pixel alignment loss $\mathcal{L}_p$. Compared to the alignments at the scene and regional levels, $\mathcal{L}_p$ serves to narrow the distance at a more intricate level, enabling more precise alignment between the two domains. Similar to the regional alignment, we begin the pixel-level alignment by computing the class probability $\mathcal{P} \in \mathcal{R}^{[HW \times n]}$ for each pixel, as in Eq.~\eqref{eq:pixel_prob}:
\begin{equation}
\label{eq:pixel_prob}
\small
\mathcal{P} = \frac{\mathbf{f}_{\mathrm{s} \rightarrow \mathrm{t}} \times \mathcal{T}^T}{\|\mathbf{f}_{\mathrm{s} \rightarrow \mathrm{t}}\| \, \|\mathcal{T}\|^T}
\end{equation}
Then, we use $\mathcal{P}$ to calculate the cross-entropy loss with the ground truth $\vec{y}\in \mathcal{R}^{[HW]}$ using Eq.~\eqref{eq:pixel_contras}:
\begin{equation}
\label{eq:pixel_contras}
\small
\mathcal{L}_p=-\frac{1}{H W} \sum_{i=1}^{HW} \vec{y}_{i} \log \left(\mathcal{P}_{i}\right)
\end{equation}

\mypara{The overall objective.}
To this end, the training objective $\mathcal{L}_{HC}$ for hierarchical context alignment is formulated as:
\begin{equation}
\small
\mathcal{L}_{HC}=\lambda_{r}\mathcal{L}_{r}+\lambda_{p}\mathcal{L}_{p}+\mathcal{L}_{\mu, \sigma}\left(\mathbf{f}_{\mathrm{s} \rightarrow \mathrm{t}}, \operatorname{TrgEmb}\right)
\end{equation}

\begin{figure}
    \centering
    \includegraphics[width=.8\linewidth]{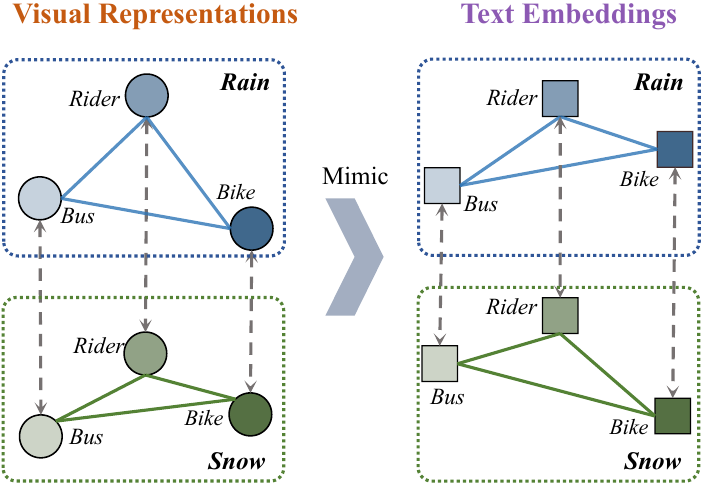}
    \caption{ \textbf{Domain Consistent Representation Learning.} We ensure the visual regional representations have similar correlations with that of text embeddings, both within the same domain and across different domains.
    }
    \label{fig:consistent}
\vspace{-0.5cm}
\end{figure}
\subsection{Domain Consistent Representation Learning}
Despite successfully bringing the feature $\mathbf{f}_{\mathrm{s} \rightarrow \mathrm{t}}$ closer to the target domain, this process may unintentionally interfere with the relational information among different prototypes. As depicted in Fig.~\ref{fig:consistent}, although the entities `rider,' `bus,' and `bike' from the source domain have been individually aligned with their target domain text embeddings, the inherent correlations between these classes might be disrupted due to the domain shift. Furthermore, the same category in different contexts may exhibit distinct correlations between the visual and text representations. 

For instance, the text embeddings of ``a bus in the snow," ``a bus in the rain," and ``a bus at night" may have different correlations compared to their visual counterparts in the contexts of `snow', `rain' and `night' respectively. 
This discrepancy in relational consistency between the simulated domain features and text embeddings can lead the model to erroneously diverge from the true distributions represented by the text embeddings of the target domain. To tackle this problem, we propose the domain consistency loss $\mathcal{L}_{DC}$.

Specifically, for $n$ categories in $m$ target domains, with Eq.~\eqref{eq:get_proto}, we can obtain $m$ prototypes $\mathcal{C} \in \mathcal{R}^{[n \times d]}$
we group it into the $\widetilde{\mathcal{C}}
\in \mathcal{R}^{[mn \times d]}$. 
Similarly, we obtain the $m$ text embedding $\mathcal{T} \in \mathcal{R}^{[n \times d]}$,
 grouped into the $\widetilde{\mathcal{T}}
\in \mathcal{R}^{[mn \times d]}$. Lastly, we adopt Eq.~\eqref{eq:similarity_domain} as $\mathcal{L}_{DC}$ to enforce representation consistency across multiple domains by preserving the correlation between the prototypes and the corresponding text embeddings in different scenes.
\begin{equation}
\label{eq:similarity_domain}
\small
\vspace{-0.05cm}
\mathcal{L}_{DC} = MSE( \frac{ \widetilde{\mathcal{C}} \widetilde{\mathcal{C}}^T}{\| \widetilde{\mathcal{C}}\|^2}, \frac{ \widetilde{\mathcal{T}} \widetilde{\mathcal{T}}^T}{\| \widetilde{\mathcal{T}}\|^2})
\end{equation}
\subsection{Text-Driven Rectifier}
\begin{figure}[!t]
    \centering
    \includegraphics[width=.45\linewidth]{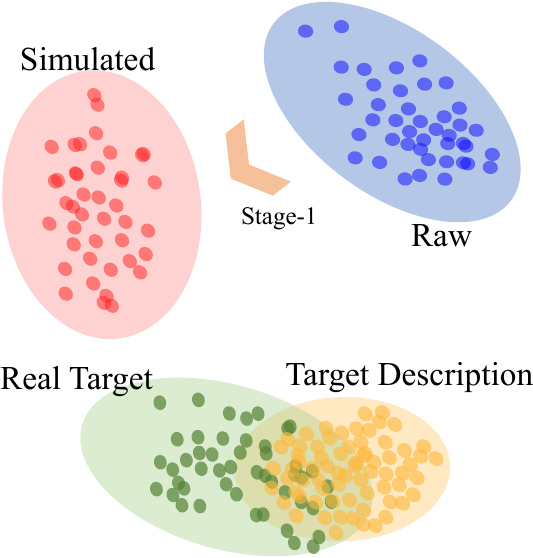}
    \caption{\textbf{Text-Driven Rectifier.} Despite the simulated features exhibiting a closer alignment with the real targets in comparison to raw features, a gap still exists. To address this disparity, we employ the text descriptions from the target domain to rectify the simulated ones when tuning the segmentation head in Stage-2. 
    }

\vspace{-0.3cm}
    \label{fig:critify}
\end{figure}

\mypara{Evils in the simulated features.} 
During the second stage introduced in Sec.~\ref{sec:preliminary}, the model utilizes the simulated target domain features to fine-tune the segmentation head, enabling the model to be effectively adapted to the target domain. 
However, as shown in Fig.~\ref{fig:critify}, discrepancies may persist between the simulated features and the actual target domain features. It is crucial to consider these discrepancies as using simulated features directly may lead to a deviation of the segmentation head from the true target distributions, yielding worse segmentation performance after tuning.

Therefore, we propose to address this issue by leveraging the text embeddings obtained from CLIP, which effectively resemble the distributions in the real target domain. By adopting these text embeddings as a prior, we may rectify the simulation process, thereby encouraging the simulated features to align more closely with the target features.

\mypara{Rectification benefits adaptation.}
Specifically, we denote the features simulated by PIN as $\mathbf{f}_{\mathrm{s} \rightarrow \mathrm{t}}$, \textit{i.e.}
$\mathbf{f}_{\mathrm{s} \rightarrow \mathrm{t}} = \operatorname{PIN}\left(\mathbf{f}_{\mathrm{s}}, \boldsymbol{\mu}, \boldsymbol{\sigma}\right)$.
Then, we get the rectified feature $\mathbf{\widetilde{f}}_{\mathrm{s} \rightarrow \mathrm{t}}$ by following Eq.~\eqref{eq:rectify}:

\begin{equation}
\label{eq:rectify}
\small
\mathbf{\widetilde{f}}_{\mathrm{s} \rightarrow \mathrm{t}} = \beta \left( \boldsymbol{\widetilde{\sigma}} \left( \frac{\mathbf{f}_{\mathrm{s} \rightarrow \mathrm{t}} - \mu\left(\mathbf{f}_{\mathrm{s} \rightarrow \mathrm{t}}\right)}{\sigma\left(\mathbf{f}_{\mathrm{s} \rightarrow \mathrm{t}}\right)} \right) + \boldsymbol{\widetilde{\mu}} \right) + \mathbf{f}_{\mathrm{s} \rightarrow \mathrm{t}}, 
\end{equation}
where $\beta$ is a learnable factor, initialized as 0.1, controlling the extent of the rectification applied to $\mathbf{f}_{\mathrm{s} \rightarrow \mathrm{t}}$. $\widetilde{\sigma}$ and $\widetilde{\mu}$ are obtained by passing text embedding through a linear layer to represent the mean and standard deviation of the target domain features, respectively. Then we utilize the $\mathbf{\widetilde{f}}_{\mathrm{s} \rightarrow \mathrm{t}}$ obtained from multiple domains to fine-tune the head.

Consequently, through text-driven rectification (TDR), we are able to correct the features initially simulated by PIN to preserve the distinctive attributes of each domain. This enhancement improves the overall generalization capability of the shared final head, enabling it to effectively adapt to multiple domains simultaneously.

It is worth noting that TDR is applied to bridge the gap between simulated features and actual target domain features, exclusively during Stage-2 mentioned in Sec.~\ref{sec:preliminary} for head tuning. We do not apply TDR to Stage-1 as it would lead to a trivial solution. The theoretical proof is in Sec.~\ref{sec:discussion}.

\subsection{Overall Loss Function}
With the above strategies, the overall loss function $\mathcal{L}$ for Stage-1's training becomes:
\begin{equation}
\mathcal{L}= \lambda_{HC}\mathcal{L}_{HC}+\lambda_{DC}\mathcal{L}_{DC}+\lambda_{seg}\mathcal{L}_{seg}
\end{equation}
where the $\lambda_{HC}$, $\lambda_{DC}$, $\lambda_{seg}$, are weighting coefficients balancing the respective loss components. We observe the segmentation loss $\mathcal{L}_{seg}$ benefits the Stage-1 training, as illustrated in the ablation study.
For Stage-2, which involves fine-tuning the model for segmentation, only the vanilla segmentation loss $\mathcal{L}_{seg}$ is adopted.

\section{Experiments}
\label{sec:Experiment}
In Sec.~\ref{sec:exp_setup}, we present the details of our experiment setup, encompassing the datasets, task settings, and implementation specifics. We showcase the effectiveness of our method in both traditional settings, demonstrated in Sec.~\ref{sec:effect_tradition}, and in new practical settings, elaborated upon in Sec.~\ref{sec:effect_new}. 
Additionally, we utilize GPT-4 to generate multiple natural descriptions of autonomous driving scenarios, including some uncommon situations such as sandstorms and forest fires. These scenarios are often challenging for model training due to their limited data availability. Due to space limitations, we detail this interesting experiment in the supplementary.

\subsection{Experiment Setup}
\label{sec:exp_setup}

\textbf{Datasets.} We primarily use the Cityscapes~\cite{cordts2016cityscapes} as the source domain dataset. 
Following P{\O}DA\xspace, we report the main results using ACDC~\cite{sakaridis2021acdc}.
To demonstrate the generalization of our method, we also investigate two extra adaptation scenarios: 
real to synthetic (source: Cityscapes; target: GTA5~\cite{richter2016playing}) and synthetic to real (source: GTA5; target: Cityscapes). The evaluation configuration follows~\cite{fahes2023poda}.

\mypara{Implentation Details}. In this study, 
we conduct comparisons on both the setting of P{\O}DA\xspace and our proposed setting.
As for the base segmentation model, DeepLabv3+~\cite{chen2018encoder} with a backbone model of pre-trained CLIP-ResNet-50\footnote{\url{https://github.com/openai/CLIP}} is adopted. The base model is sufficiently trained on the source domain following the configuration of~\cite{fahes2023poda}.
In the fine-tuning stage~(Stage 2), we begin with the source pre-trained model and only fine-tune the classifier head, also following the configurations of~\cite{fahes2023poda} for a fair comparison. 
All models are tested on the original images without resizing, and more details are in the supplementary file.

\subsection{Effectiveness on Traditional Settings.}
\label{sec:effect_tradition}
\mypara{Effectiveness on prompt-driven zero shot adaptation.}
Following the previous benchmark, we explore various adaptation scenarios, including: \DAsetting{day}{night}, \DAsetting{clear}{snow}, \DAsetting{clear}{rain}, \DAsetting{real}{synthetic}, and \DAsetting{synthetic}{real}. We compare our approach with two state-of-the-art baselines: CLIPstyler~\cite{kwon2022clipstyler} for zero-shot style transfer and {P{\O}DA\xspace}~\cite{fahes2023poda} for prompt-driven zero-shot adaptation. Notably, {P{\O}DA\xspace}, CLIPstyler and our approach, do not utilize target images during training. Following the previous setting, we only select simple prompts for each domain to demonstrate the effectiveness of our method. 

\begin{table}[t!]
		\setlength{\tabcolsep}{0.05\linewidth}
		\centering
		\resizebox{0.9\linewidth}{!}{
                \newcolumntype{H}{>{\setbox0=\hbox\bgroup}c<{\egroup}@{}}
    \begin{tabular}{lllc}
			\toprule
			Source
			& Target eval.
			& Method
			& mIoU[\%]\\
			\midrule
        \multirow{20}{*}{CS}& \multicolumn{3}{c}{\cellcolor{gray!34}Prompt = ``driving at night''}\\
			& \multirow{4}{*}{ACDC Night} & source-only & 18.31 \\
    & & CLIPstyler & 21.38\\
    & & {P{\O}DA\xspace} & 25.03 \\
    & & \ours & \textbf{25.40} \\

        & \multicolumn{3}{c}{\cellcolor{gray!34}Prompt = ``driving in snow''}\\
        & \multirow{4}{*}{ACDC Snow} & source-only &  39.28 \\
    & & CLIPstyler & 41.09 \\
        & & {P{\O}DA\xspace} & 43.90 \\
          & & \ours & \textbf{46.00} \\

        &\multicolumn{3}{c}{\cellcolor{gray!34}Prompt = ``driving under rain''}\\
		& \multirow{4}{*}{ACDC Rain} & source-only & 38.20 \\
  & & CLIPstyler & 37.17 \\
  & & {P{\O}DA\xspace} &  42.31\\
  & & \ours &  \textbf{44.94}\\

    &\multicolumn{3}{c}{\cellcolor{gray!34}Prompt = ``driving in a game''}\\
    & \multirow{4}{*}{GTA5} & source-only & 39.59 \\
    & & CLIPstyler & 38.73 \\
    & & {P{\O}DA\xspace} & 40.77 \\
    & & \ours & \textbf{42.91} \\

	\arrayrulecolor{black}		
	\midrule
	\multirow{5}{*}{GTA5} &\multicolumn{3}{c}{\cellcolor{gray!34}Prompt = ``driving in real''}\\
    & \multirow{4}{*}{Cityscapes} & source-only &  36.38 \\
    & & CLIPstyler &  32.40 \\
    & & {P{\O}DA\xspace} & 40.02 \\
     & & \ours & \textbf{41.73} \\
    \bottomrule
    \end{tabular}}
    \caption{\textbf{Performance on classic prompt driven zero shot domain adaptation in semantic segmentation.} Performance (mIoU\%) of~\ours~framework compared against previous methods and source-only baseline. CS stands for Cityscapes~\cite{cordts2016cityscapes}. }
	\label{tab:main_results}
 \vspace{-0.4cm}
\end{table}

As shown in Table~\ref{tab:main_results}, our proposed method consistently outperforms all baseline models in zero-shot domain adaptation, using mean Intersection over Union (mIoU) as the comparative metric. Our method surpasses previous approaches, achieving improvements in all the scenarios.
It is noteworthy that the previous SOTA method, P{\O}DA\xspace, requires training a separate head for each scenario. We believe that while using distinct heads for individual scenarios simplifies the task, it also compromises the method's generalizability, limiting its practical application in real-world settings. In contrast, our method surpasses previous approaches by using only a single head, further demonstrating our method's effectiveness.
Furthermore, our proposed method, without altering the original framework or requiring any additional information, achieves significant improvements by deeply exploring the relationships between multi-level images and texts. This further validates the effectiveness of our approach in the traditional zero-shot domain adaptation task.

Besides, in supplementary, we compare our proposed method with the one-shot SOTA method SM-PPM~\cite{wu2022style} to demonstrate the effectiveness of our method.

\begin{table*}[htb]
\centering
\setlength\tabcolsep{11pt}
\begin{adjustbox}{width=1\linewidth,center=\linewidth}
\begin{tabular}{c|c|cc|cc|cc|cc|c }
\hline

\multicolumn{2}{c|}{Scenarios}          & \multicolumn{2}{c|}{Source2Fog}    & \multicolumn{2}{c|}{Source2Night}     & \multicolumn{2}{c|}{Source2Rain}  & \multicolumn{2}{c|}{Source2Snow}    & \multirow{3}{*}{Mean-mIoU}  \\ \cline{1-10}
\multicolumn{2}{c|}{Domain Description}          & \multicolumn{2}{c|}{driving in fog}    & \multicolumn{2}{c|}{driving at night}     & \multicolumn{2}{c|}{driving under rain}  & \multicolumn{2}{c|}{driving in snow}    & \\ \cline{1-10}
Method & REF & mIoU  & mAcc  & mIoU  & mAcc  & mIoU  & mAcc  & mIoU  & mAcc  & \\ \hline

Source & OpenAI&49.98&65.42&18.31&34.16&38.20&58.97&39.28&54.64&36.44\\ 
CLIPStyler  & CVPR2022 \cite{kwon2021clipstyler}  &48.87&64.31&  20.83&35.32&  36.97&57.46&  40.31&54.42 &36.75\\ 
$\text{P{\O}DA\xspace}^*$& ICCV2023 \cite{fahes2023poda} &51.54&64.51 &25.03&55.5 &42.31&75.4 &43.90&70.7&40.65\\ 
\ours  & ours&\textbf{53.55}&\textbf{80.2}&\textbf{25.40}&\textbf{55.8}&\textbf{44.94}&\textbf{74.4}&\textbf{46.00}&\textbf{70.0}&\textbf{42.47}\\ 

 \hline
\end{tabular}
\end{adjustbox}

\caption{\textbf{Performance comparison of clear-to-adverse weather in ULDA.} We use Cityscape as the source domain and ACDC as the four target domains in this setting. Mean-mIoU represents the average mIoU value in four scenarios. $\text{P{\O}DA\xspace}^*$ represents the model that uses different segmentation heads in specific domains, while the others adopt the shared head.  }
\label{tab:ULDA}
\end{table*}

\begin{table*}[htb]

\vspace{-0.3cm}
\centering
\setlength\tabcolsep{11pt}
\begin{adjustbox}{width=1\linewidth,center=\linewidth}
\begin{tabular}{c|cc|cc|cc|cc|cc|c }
\hline

\multicolumn{1}{c|}{ Scenarios}   & \multicolumn{2}{c|}{Source2CS}       & \multicolumn{2}{c|}{Source2Fog}    & \multicolumn{2}{c|}{Source2Night}     & \multicolumn{2}{c|}{Source2Rain}  & \multicolumn{2}{c|}{Source2Snow}    & \multirow{3}{*}{Mean-mIoU}  \\ \cline{1-11}
\multicolumn{1}{c|}{Domain Description}    & \multicolumn{2}{c|}{driving in real}      & \multicolumn{2}{c|}{driving in fog}    & \multicolumn{2}{c|}{driving at night}     & \multicolumn{2}{c|}{driving under rain}  & \multicolumn{2}{c|}{driving in snow}    & \\ \cline{1-11}
Method &  mIoU  & mAcc & mIoU  & mAcc  & mIoU  & mAcc  & mIoU  & mAcc  & mIoU  & mAcc  & \\ \hline

Source & 36.38&46.19&33.20&42.51&12.22&22.56&33.32&43.15&32.33&40.60&29.49\\ 
CLIPStyler  &32.20&41.64  &30.79&40.37&  11.12&20.18&31.17&  40.06&30.65 &38.97 &27.19\\ 
$\text{P{\O}DA\xspace}^*$& 40.05&48.95&35.76&44.98 &13.35&25.24 &34.19&45.93 &33.81&42.10&31.43\\ 
\ours  & \textbf{41.73}&\textbf{51.98}&\textbf{36.98}&\textbf{46.56}&\textbf{15.72}&\textbf{28.99}&\textbf{35.84}&\textbf{47.39}&\textbf{35.77}&\textbf{43.74}&\textbf{33.21}\\ 

 \hline
\end{tabular}
\end{adjustbox}

\caption{\textbf{Performance comparison of Synthetic-to-Real in ULDA} We use GTA5 as the source domain, Cityscapes and ACDC as the five target domains in this setting. Mean-mIoU represents the average mIoU value in five scenarios. $\text{P{\O}DA\xspace}^*$ represents the model used different segmentation heads in specific domains, while the others adopt the shared head. }
\label{tab:GTA}
\vspace{-0.3cm}
\end{table*}
\subsection{Effectiveness on Unified language-driven zero shot adaptation}
\label{sec:effect_new}
\mypara{Quantitative Results.}
For the newly introduced task of Unified Language-driven Zero-Shot Adaptation, our aim is to propose a benchmark that is more aligned with real-world scenarios and holds practical value. Accordingly, following the setting described in Sec.~\ref{Sec: task settings}, we are limited to only having natural descriptions of potential target domains. And this setting also requires our model to be versatile enough to be tested across all downstream target domains using just a singular model architecture.
We set two practical adaptation scenarios as our benchmark, including: clear-to-adverse-weather adaptation on Cityscapes$\rightarrow$ACDC and synthetic-to-real adaptation on GTA5$\rightarrow$Cityscape and ACDC.

We establish the mean mIoU as the comparative metric for our study. This mean mIoU is derived by calculating the average of mIoU values across various domains. Additionally, we also report the mIoU and mean Accuracy (mAcc) for each individual domain. To demonstrate the generalizability of our method, we just utilize simple prompt descriptions to reflect the target domain knowledge.

\textbf{Clear-to-Adverse weather.} In Table~\ref{tab:ULDA}, we compare our proposed \ours~with the SOTA method in Zero-shot domain adaptation. $\text{P{\O}DA\xspace}^*$ employs four distinct heads, selecting a specific head tailored to each scenario.  
Our proposed method consistently surpasses all previous models in Unified Language-driven Domain Adaptation. It achieves improvements of 6.03\% over the baseline source model.
Remarkably, our approach, which utilizes a single head, even exceeds the performance of $\text{P{\O}DA\xspace}^*$, and achieves improvements of 1.82\% mIoU, which necessitates training separate heads for different scenarios. 
This highlights the strength of our method, particularly in its ability to employ Hierarchical Context modeling. Such modeling adeptly extracts and leverages the multi-level correlation between image and text, facilitating more effective domain transfer. Moreover, our approach of domain-consistent representation learning ensures consistency across various domains. This enables our model to generalize effectively to various domains within a single unified model architecture.

\mypara{Synthetic-to-Real.} 
In traditional settings, the capability of methods in synthetic-to-real scenarios is typically validated first on the GTA5$\rightarrow$Cityscapes, and then further verified through additional experiments to demonstrate their applicability to adverse weather conditions, such as adaptation from Cityscapes to ACDC. 
However, the practical utility and generalizability of these experiments are somewhat limited. In autonomous driving scenarios, it is more desirable to learn from a diverse range of source domain virtual datasets like GTA5 and to achieve direct generalization to various complex real-world scenarios, like ACDC.
Therefore, our experimental setting, termed "Synthetic-to-Real," specifically focuses on the GTA5$\rightarrow$Cityscapes$+$ACDC dataset.

In Table~\ref{tab:GTA}, we present a comprehensive comparison of our proposed \ours~method against the current SOTA methods in Zero-shot domain adaptation. Our method consistently outperforms all previous models in Unified Language-driven Domain Adaptation, achieving significant improvements of 3.72\% and 1.78\% over the baseline source model and the former SOTA method, $\text{P{\O}DA\xspace}^*$, respectively.
Notably, our method shows remarkable performance enhancement in complex scenarios. For instance, in GTA5-to-Cityscapes where the source model only achieves 36.38\% mIoU and 46.19\% mAcc, our proposed method achieves a 5.35\% mIoU and 5.79\% mAcc increase. This highlights that our method can effectively learn extensive knowledge directly from GTA5 through language alone and generalize to real and complex scenarios like Cityscapes ACDC. Such results underscore the effectiveness and practicality of our approach. It demonstrates our method's capability to extract and apply knowledge from the language description of the target domains, proving its utility and adaptability in real-world complex environments.

\begin{figure}[!t]
    \centering
    \includegraphics[width=1.0\linewidth]{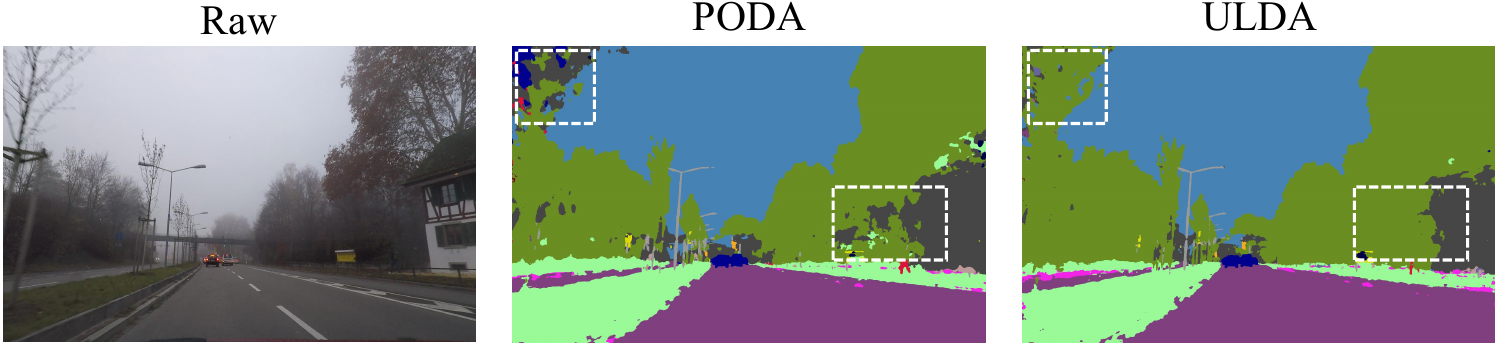}
    \caption{\textbf{Qualitative Results}  of our models and previous SOTA method on ACDC-Foggy. More results are in the Appendix.
    }
    \label{fig:visualize}
    \vspace{-0.5cm}
\end{figure}

\mypara{Qualitative Results}
To demonstrate the effectiveness of our proposed method, we show the qualitative analysis, shown in Fig.~\ref{fig:visualize}. Other visualization results are shown in the supplementary.

\subsection{Effectiveness of each component}

We conduct the ablation study on Cityscapes-to-ACDC in the Language-driven zero shot adaptation setting and evaluate the contribution of each component in our method, including Hierarchical Context Alignment~(HCA), Domain Consistent Representation Learning~(DCRL) and Text-Driven Domain Rectify~(TDR). Mean-mIoU and Mean-mAcc are used as metrics, representing the average mIoU and mAcc values across four scenarios, respectively.

\mypara{Effects of HCA and DCRL.}
As shown in Table~\ref{tab:ablation}, $Ex_1$ represents the baseline method, {P{\O}DA\xspace}, which only leverages the scene-level alignment. In comparison, $Ex_2$ introduces HCA to align image embeddings and natural descriptions on a multi-level basis. This approach results in improvements of 0.85\% in mean-mIoU.
In $Ex_3$, the integration of DCRL leads to further enhancements, with an increase of 0.97\% in mean-mIoU. This signifies that incorporating Domain Consistent Representation Learning effectively addresses domain discrepancies. For a detailed examination of the impact of DCRL, refer to the further ablation study provided in the Appendix.

\mypara{Effects of TDR and segmentation loss.}
Building on $Ex_2$, $Ex_4$ introduces Text-Driven Domain Rectify (TDR). TDR, by making rectify during the fine-tuning phase, bridges the gap between the simulated features and the real features of the target domain. This results in a performance improvement of 0.83\% in mean-mIoU.
In $Ex_5$, based on the foundation laid by $Ex_3$, we introduce a downstream task loss: $\mathcal{L}_{seg}$, during the fine-tuning phase. This loss function helps prevent the overfitting of features to text embeddings and ensures the retention of downstream task capabilities. It effectively maintains a balance between domain adaptation and task-specific performance. By incorporating this element, $Ex_5$ achieves an additional performance improvement of 0.33\% in mean-mIoU.

Overall, $Ex_{6}$ shows the complete combination of all components, achieving 42.47\% mean-mIoU in total. This demonstrates that all components compensate each other and jointly address the challenge in Language-driven zero-shot domain adaptation.
Due to the limitation of the space, the additional ablation studies are shown in the Appendix.

\begin{table}[!tb]
\vspace{-0.3cm}
\centering
\setlength\tabcolsep{4.0pt}
\renewcommand\arraystretch{1}
\begin{tabular}{c|cccc|c}
\toprule
 & \makecell*[c]{HCA} & \makecell*[c]{DCRL} & \makecell*[c]{TDR} & \makecell*[c]{$\mathcal{L}_{seg}$}  & Mean-mIoU \\\midrule
$Ex_{1}$ &  & & & &39.67 \\ 
$Ex_{2}$& \checkmark &  & &  & 40.52\\
$Ex_{3}$ &\checkmark  & \checkmark &  & & 41.49\\
$Ex_{4}$  & \checkmark &  &\checkmark  &  &41.35\\
$Ex_{5}$  & \checkmark & \checkmark & &\checkmark & 41.68\\
$Ex_{6}$ & \checkmark &  \checkmark &\checkmark  &\checkmark & 42.47\\
\bottomrule
\end{tabular}
\caption{\textbf{Ablation: Contribution of each component. 
}}
\label{tab:ablation}
\vspace{-0.42cm}

\end{table}

\section{Further Discussions}
\label{sec:discussion}
\textbf{Why not incorporate TDR to Stage-1?}
For the original source domain feature, we can obtain the corresponding target domain feature $\mathbf{f}_{\mathrm{s} \rightarrow \mathrm{t}}$ through the following formula:
\begin{equation*}
\small
\mathbf{f}_{\mathrm{s} \rightarrow \mathrm{t}}=\operatorname{PIN}\left(\mathbf{f}_{\mathrm{s}}, \boldsymbol{\mu}, \boldsymbol{\sigma}\right)=\boldsymbol{\sigma}\left(\frac{\mathbf{f}_{\mathrm{s}}-\mu\left(\mathbf{f}_{\mathrm{s}}\right)}{\sigma\left(\mathbf{f}_{\mathrm{s}}\right)}\right)+\boldsymbol{\mu}.
\end{equation*}
For the simulated $\mathbf{f}_{\mathrm{s} \rightarrow \mathrm{t}}$, we have $std(\mathbf{f}_{\mathrm{s} \rightarrow \mathrm{t}})= \sigma$, $mean(\mathbf{f}_{\mathrm{s} \rightarrow \mathrm{t}})= \mu$. Substituting them into Eq.~\eqref{eq:rectify} yields:
\begin{equation}
\small
\vspace{0.5cm}
\begin{aligned}
 \mathbf{\widetilde{f}}_{\mathrm{s} \rightarrow \mathrm{t}}&=\beta \left( \boldsymbol{\widetilde{\sigma}} \left( \frac{\mathbf{f}_{\mathrm{s} \rightarrow \mathrm{t}} - \mu\left(\mathbf{f}_{\mathrm{s} \rightarrow \mathrm{t}}\right)}{\sigma\left(\mathbf{f}_{\mathrm{s} \rightarrow \mathrm{t}}\right)} \right) + \boldsymbol{\widetilde{\mu}} \right) + \mathbf{f}_{\mathrm{s} \rightarrow \mathrm{t}} \\
& =\beta\left(\boldsymbol{\widetilde{\sigma}} \left( \frac{\mathbf{f}_{\mathrm{s}}-\mu\left(\mathbf{f}_s\right)}{\sigma\left(\mathbf{f}_{\mathrm{s}}\right)} \right) +\boldsymbol{\widetilde{\mu}}\right)+u+\sigma\left(\frac{\mathbf{f}_{\mathrm{s}}-\mu\left(\mathbf{f}_{\mathrm{s}}\right)}{\sigma\left(\mathbf{f}_{\mathrm{s}}\right)}\right) . \\
& =\left( \frac{\mathbf{f}_{\mathrm{s}}-\mu\left(\mathbf{f}_{\mathrm{s}}\right)}{\sigma\left(\mathbf{f}_{\mathrm{s}}\right)} \right) (\beta \boldsymbol{\widetilde{\sigma}}+\sigma)+(\beta\boldsymbol{\widetilde{\mu}}+u) . \\
\end{aligned}
\vspace{-0.5cm}
\end{equation}
$\boldsymbol{\widetilde{\sigma}}$ and $\boldsymbol{\widetilde{\mu}}$ are derived by passing text embeddings through a linear layer. The parameters $\sigma$ and $\mu$ are learnable and are designed to simulate features of the target domain. During Stage-1, it is necessary to optimize $\mu$ and $\sigma$ to transform the source domain features into those of the target domain, ensuring alignment with the text embeddings. However, as the text embeddings are directly input into the linear layer to obtain $\widetilde{\mu}$ and $\widetilde{\sigma}$, this process results in $\mu$ and $\sigma$ not being optimized, leading to a trivial solution. Therefore, we may not integrate rectification into Stage-1. A more detailed simplification process is shown in the Appendix. 

\mypara{Is ULDA a degraded form of domain generalization?}
Our propose ULDA is not a degraded form of domain generalization~(DG)~\cite{wang2022generalizing}. Because the ULDA does not conflict with DG; rather, it complements them. DG methods typically utilize meta-learning or feature alignment techniques to incorporate domain-invariant information from source data during the training phase within the source domain. In contrast, ULDA focuses on enhancing a pre-trained model, enabling it to generalize more efficiently and effectively across a wider range of target domains. Subsequent experiments in the supplementary demonstrate that our proposed method can also yield benefits for DG methods.

\section{Concluding Remarks}
This work spots issues in the literature and presents a new setting named Unified Language-driven Zero-shot Domain Adaptation (ULDA) with three simple yet effective strategies Hierarchical Context Alignment (HCA), Domain Consistent Representation Learning (DCRL), and Text-Driven Rectifier (TDR). The effectiveness and practical merits of our method have been verified by the decent performance achieved by challenging benchmarks without imposing any additional inference burdens. 

\textbf{Acknowledgements}. This work receives partial support from the Shenzhen Science and Technology Program under  No. KQTD20210811090149095. 
\clearpage
\newpage
{
    \small
    \bibliographystyle{ieeenat_fullname}
    \bibliography{main}
}
\clearpage
\newpage
\appendix
\section*{Overview}

\begin{itemize}[itemsep=1pt, parsep=5pt]
    \item ULDA on Autonomous Driving
    \begin{itemize}
        \item Motivation~\ref{sec:motivation}
        \item Proposed Adverse Weather Benchmark~\ref{sec:benchmark}
        \item Results~\ref{sec:results}
    \end{itemize}
    \item Additional Experiments

    \begin{itemize}
        \item Effectiveness Compared One-shot Domain Adaptation~\ref{sec:osda}
        \item Effectiveness on Domain Generalization~\ref{sec:dg}
        \item Additional Ablation Study~\ref{sec:ablation}
    \end{itemize}
    \item Implementation Details~\ref{sec:details}
    \item Additional Discussions~\ref{sec:discussion}
    \item Related Work~\ref{sec:related}
    \item Qualitative Analysis~\ref{sec:qualitative}
\end{itemize}
\section{ULDA on Autonomous Driving}
\label{sec:drive}
\subsection{Motivation}
\label{sec:motivation}
Autonomous driving has wide applications for intelligent transportation systems, such as reducing the labor
costs, enhancing the comfortableness of customers, and so on~\cite{xu2021opencda, jo2014development}.
In some cases, the autonomous vehicle
might work in adverse weather conditions, like night, rain, fog, and so on.
These complex scenarios might take a big challenge to the autonomous driving system.

However, in practical terms, it is not always feasible to obtain comprehensive data for every possible adverse condition due to the high costs and difficulties associated with data collection. 
Instead, practitioners may only have a conceptual understanding or hypothetical descriptions of potential driving scenarios. 
In this case, the ability to augment a model's performance in such predicted scenarios without actual data collection is preferred. 

Therefore, our proposed Unified Language-driven Zero-shot Domain Adaptation~(ULDA) holds great potential in autonomous driving scenarios and could substantially enhance the future advancement of autonomous driving technology.

\begin{table*}[htb]
\centering
\setlength\tabcolsep{11pt}
\renewcommand{\arraystretch}{1.1} 
\Huge
\begin{adjustbox}{width=1\linewidth,center=\linewidth}
\begin{tabular}{c|c|c|c|c|c|c|c }
\hline

\multicolumn{1}{c|}{Scenarios} & \multicolumn{1}{c|}{Source2Fog} & \multicolumn{1}{c|}{Source2Night} & \multicolumn{1}{c|}{Source2Rain} & \multicolumn{1}{c|}{Source2Snow} & \multicolumn{1}{c|}{Source2Sandstorm} & \multicolumn{1}{c|}{Source2Fire} &\multirow{3}{*}{mean-mIoU} \\ \cline{1-7}
\multicolumn{1}{c|}{\multirow{2}{*}{Domain Description}} & \multicolumn{1}{c|}{\multirow{2}{*}{driving in fog}} & \multicolumn{1}{c|}{\multirow{2}{*}{driving at night}} & \multicolumn{1}{c|}{\multirow{2}{*}{driving under rain}} & \multicolumn{1}{c|}{\multirow{2}{*}{driving in snow}} & \multicolumn{1}{c|}{\multirow{2}{*}{driving in sandstorm}} & \multicolumn{1}{c|}{\multirow{2}{*}{driving through fire}} \\ 
& & & & & & \\ \cline{1-8}

Source & 49.98 & 18.31 & 38.20 & 39.28 & 19.58 & 10.08 &27.57\\ 
CLIPStyler & 48.87 & 20.83 & 36.97 & 40.31 & 23.16 & 12.36 &30.42 \\ 
$\text{P{\O}DA\xspace}^*$ & 51.54 & \textbf{25.03} & 42.31 & 43.90 & 24.39 & 15.43 &33.77 \\ 
\ours & \textbf{53.02} & 24.61 & \textbf{45.12} & \textbf{46.06} & \textbf{25.72} & \textbf{19.52} &\textbf{35.68} \\ 

\hline
\end{tabular}
\end{adjustbox}

\caption{\textbf{Performance of ULDA on Autonomous Driving.} We use Cityscape as the source domain and ACDC and our collected data as the six target domains in this setting. Mean-mIoU represents the average mIoU value in six scenarios. $\text{P{\O}DA\xspace}^*$ represents the model that uses different segmentation heads in specific domains with domain-id provided, while our ULDA utilizes a unified, all-in-one head.  }
\label{tab:Driving}
\end{table*}

\subsection{Proposed Benchmark}
\label{sec:benchmark}
To further explore the potentially challenging autonomous driving scenarios, we use GPT-4 to generate several difficult situations. Our prompt is ``Please describe some potential adverse driving scenarios, which pose challenges for autonomous driving, such as `Driving in fog', `Driving in snow'.''
After careful consideration of the comprehensive answers provided by GPT-4 and the scene understanding capabilities of CLIP, we choose `sandstorm' and `fire' as additional scenarios to augment the existing autonomous driving scenes (rain, snow, fog, night). These scenarios have been selected based on their challenging nature and the relatively limited availability of relevant data.

Due to the rarity of these scenarios, collecting images for them has been challenging.  We make significant efforts to gather several images that fulfill the requirements for autonomous driving from publicly accessible websites with copyright permissions. These images have been annotated and will serve as a new benchmark. All collected data will be released.
\subsection{Results}
\label{sec:results}
We choose Cityscapes~\cite{cordts2016cityscapes} as the source domain and ACDC~\cite{sakaridis2021acdc} with Fog, Night, Rain, Snow, as well as our collected Sandstorm and Fire as the target domain. All other model implementation details remain consistent with the main experiments. In order to demonstrate the effectiveness of our approach, we train the state-of-the-art (SOTA) method, P{\O}DA\xspace~\cite{fahes2023poda}, with six distinct segmentation heads for specific domains. In contrast, our model is exclusively trained with a unified, all-in-one head.

As illustrated in Table~\ref{tab:Driving}, our proposed method consistently outperforms all previous models in the Autonomous Driving scenario. Our method achieves improvements of 8.11\% and 1.91\% over the baseline source model and the former state-of-the-art method, $\text{P{\O}DA\xspace}^*$, respectively. 
Notably, our approach, employing a single head, even surpasses the performance of $\text{P{\O}DA\xspace}^*$ and achieves a 1.82\% mIoU improvement, which requires training separate heads for different scenarios. 
Furthermore, in challenging scenarios like `driving through fire' and `driving under snow,' the backgrounds are almost red and white, respectively. 
Our method demonstrates a significant improvement compared to the previous approach, increasing by 4.09\% and 2.16\%, respectively.
These results demonstrate that contributed to Hierarchical Context modeling and Text-Driven Rectifier, our method can adeptly and precisely extract and utilize the multi-level correlation between images and text, thereby achieving significant improvement in complex scenarios.
Additionally, our proposed domain-consistent representation learning ensures consistency across diverse domains, enabling our model to generalize effectively under a single unified segmentation head.
What's more, we present the qualitative analysis in Sec.~\ref{sec:qualitative}.
\section{Additional Experiments}
In this section, we demonstrate the effectiveness of our method by comparing it to the One-shot Domain Adaptation method in Sec.~\ref{sec:osda}. Furthermore, we demonstrate that our method can achieve further improvement based on the domain generalization method in Sec.~\ref{sec:dg}.

\subsection{Effectiveness Compared One-shot Domain Adaptation}
\label{sec:osda}
To show the effectiveness of our~\ours, We evaluate it against SM-PPM \cite{wu2022style}\footnote{We use official code \url{https://github.com/W-zx-Y/SM-PPM}}, a SOTA method in one-shot unsupervised domain adaptation~(OSUDA). 
The OSUDA setting allows access to a single unlabeled target domain image for adapting the model to the new target domain. In SM-PPM, this image acts as an anchor for mining target styles.
For a robust comparison, we adhere to the previous settings. We employ five randomly selected target images to train the SM-PPM. Additionally, we train five different models for each image, each initialized with a unique random seed. The mean Intersection over Union (mIoU) values reported represent the average across these 25 models.
It's important to note that a direct comparison of the absolute results between the two models may not be entirely fair due to the differences in their backbones (ResNet-101 in SM-PPM versus ResNet-50 in \ours) and segmentation frameworks (DeepLabv2 in SM-PPM versus DeepLabv3+ in \ours). Therefore, our analysis focuses on the improvement each method offers over its respective naive source-only baseline, also considering the baseline's performance.
As shown in Table~\ref{tab:comparison_with_OSUDA}, in the \DAsetting{Cityscapes}{ACDC} scenario, both the absolute and relative improvements of \ours~over its source-only version surpass those of SM-PPM. 

Notably, the SM-PPM utilizes three different images to adapt three domain-specific models to the three scenarios on ACDC. However, our~\ours~does not have access to any target domain image and utilizes one unified model to adapt to any scenario.

\begin{table}[t]

    		\setlength{\tabcolsep}{0.01\linewidth}
    \centering
  \resizebox{1.\linewidth}{!}{
  \begin{tabular}{clcc}
    \toprule
    Source & Target eval. & One-shot SM-PPM\,\cite{wu2022style} & Zero-shot \ours \\
    \midrule
    \multirow{3}{*}{CS}
    & ACDC Night & 13.07\,/\,14.60~{\small($\Delta$=1.53)} & 18.31\,/\,\textbf{25.40} ~{\small({$\Delta$=7.09})}\\
    & ACDC Snow & 32.60\,/\,35.61 ~{\small({$\Delta$=3.01})} & 39.28\,/\,\textbf{46.00} ~{\small($\Delta$=6.72)}\\
    & ACDC Rain & 29.78\,/\,32.23 ~{\small($\Delta$=2.45)} & 38.20\,/\,\textbf{44.94} ~{\small($\Delta$=6.74)}\\
    \midrule
    GTA5 & CS & 36.60\,/\,42.80 ~{\small($\Delta$=6.20)} & 36.38\,/\,\textbf{42.91} ~{\small($\Delta$=6.53)} \\
    \bottomrule
  \end{tabular}}
  \caption{\textbf{Effectiveness compared to OSUDA.} 
  	Semantic segmentation performance (mIoU\%) for source\,/\,adapted models, and gain provided by adaptation ($\Delta$ in mIoU). For adaptation, SM-PPM has access to one target image and adapts three specific models on ACDC, while \ours~not has access to any target domain image and utilizes one unified model.} 
  \label{tab:comparison_with_OSUDA}
\end{table}

\begin{table}[t]
\begin{adjustbox}{width=0.9\linewidth, center}
  \begin{tabular}{llllll}
    \toprule
    Method & Fog & Night & Snow & Rain & Mean\\
    \midrule
    Source &49.98 &18.31 & 39.28 & 38.20 & 36.44\\ 
    Source-G  &51.48 &21.07 & 42.84 & 42.38 &39.69 \\
    $\text{P{\O}DA\xspace}^*$ & 51.54 &25.03 & 42.31 &43.90 & 40.65\\ 
    $\text{P{\O}DA\xspace}^*$-G &52.87 &24.86 & 44.34 & 43.17 &41.31 \\
    \ours &53.55 & 25.40& 44.94 & 46.00 & 42.47 \\
    \ours-G &54.21 & 25.94& 46.02 & 47.15 & 43.33  \\
    \bottomrule
  \end{tabular}
\end{adjustbox}
\smallskip
\caption{\textbf{Effectiveness with DG method.} `-G' means the source model is trained with the domain generalization method~\cite{fan2023towards}. Source-only-G model is enhanced with a domain generalization technique.  $\text{P{\O}DA\xspace}^*$ represents the model that uses different segmentation heads in specific domains with domain-id provided.
}
\label{tab:dg}
\end{table}

\subsection{Effectiveness on Domain Generalization}
\label{sec:dg}
Domain generalization~(DG) is a setting that aims to develop robust models that can generalize well to new, unseen domains. 
\cite{fan2023towards} is a current SOTA method in Domain generalization by simply perturbing feature channel statistics. 
Therefore, we aim to demonstrate that by incorporating the DG method, our approach can achieve further improvement. We showcase our effectiveness across four target domain scenarios in Cityscapes-ACDC. The detailed setting is followed by $\text{P{\O}DA\xspace}$~\cite{fahes2023poda} Table 7.

 First, we follow the DG sota method~\cite{fan2023towards} to train the Source-G model, which augment features by shifting the per-channel $(\boldsymbol{\mu},\boldsymbol{\sigma})$ with Gaussian noises sampled for each batch of features.
As shown in Table~\ref{tab:dg}, the Source-only-G consistently outperforms
the Source-only model, demonstrating a generalization capability in the Semantic Segmentation task. 
Moreover, when integrating the Domain Generalization technique into our proposed ULDA, significant enhancements are observed across all target domains, leading to further improvements in the mean-mIoU metric.
Notably, in comparison to the $\text{P{\O}DA\xspace}^*$-G method, our ULDA-G approach outperforms it with a mean-mIoU improvement of 2.02\%. Moreover, our ULDA method eliminates the need for Domain-ID and utilizes a single model for all scenarios.

These experiments on Domain Generalization also illustrate that our proposed ULDA setting is not a degraded version of domain generalization. Additionally, these two settings can mutually benefit each other, as discussed in Sec. 6 of the main paper.

\subsection{Additional Ablation Study}
\label{sec:ablation}
\begin{figure}
    \centering
    \includegraphics[width=1.0\linewidth]{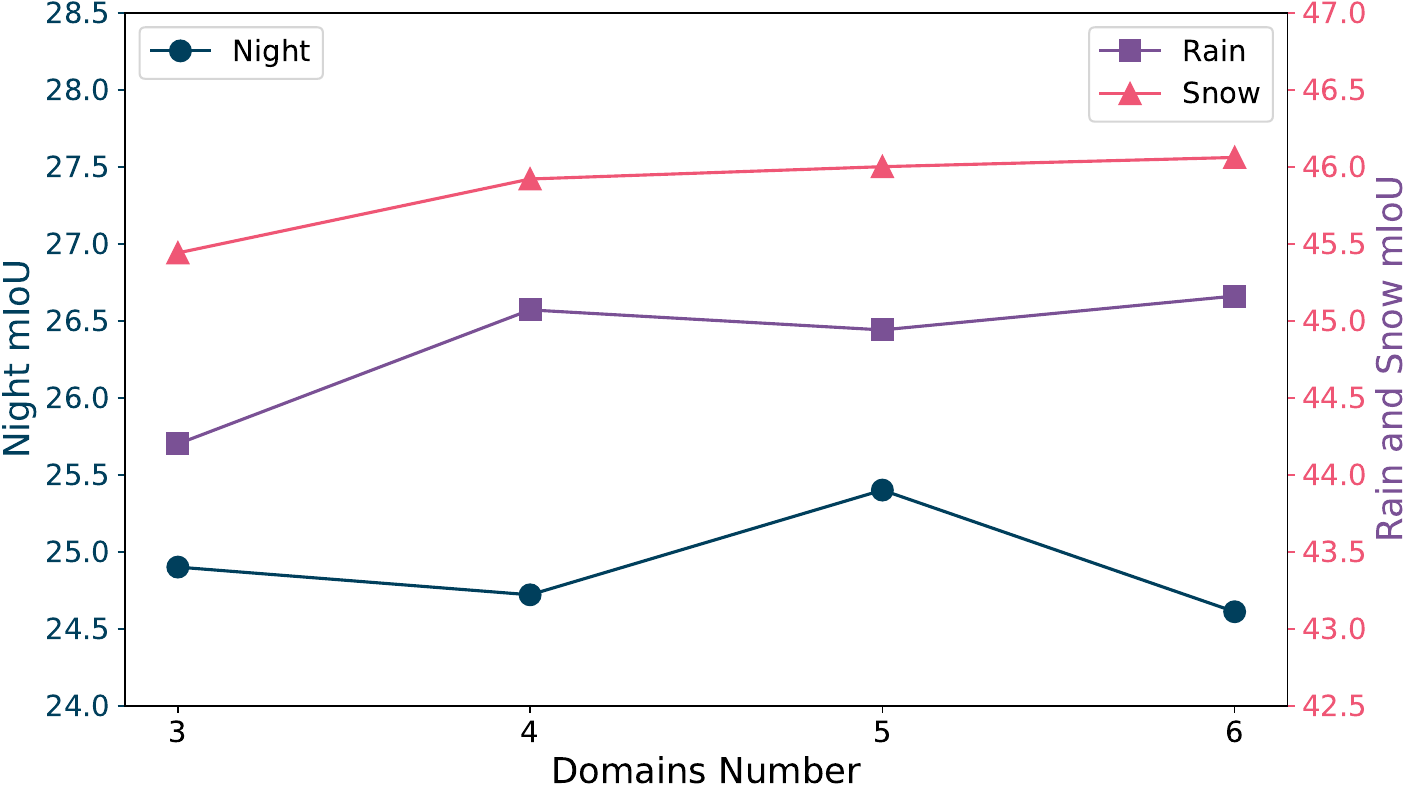}
    \caption{ \textbf{Ablation Study.} The impact of the domain numbers on the model's performance.
    }
    \label{fig:compare_num}
\end{figure}
Our proposed ULDA approach employs a single unified all-in-one segmentation head to adapt to various target domains. Consequently, a natural question arises: \textit{how does the number of domains impact the performance of the model?}
Therefore, we conduct a comparative analysis of the changes in mIoU across three scenarios, namely Night, Rain, and Snow, while varying the number of domains from 3 to 6.
Specifically, the four-domain setting includes Night, Rain, Snow, and the addition of Fog~(ACDC); the five-domain setting comprises Night, Rain, Snow, the addition of Fog~(ACDC), and GTA5; the six-domain setting involves Night, Rain, Snow, the addition of Fog~(ACDC), and our collected Sandstorm and Fire.

As shown in Fig.~\ref{fig:compare_num}, the mIoU for rainy and snowy scenes exhibits an upward trend with an increase in the number of domains. And nighttime scenes exhibit a fluctuating pattern.
This result demonstrates that our proposed Domain Consistent Representation Learning plays a crucial role in maintaining consistent performance across domains, preventing any decline in performance.
The gradual improvement in mIoU for rainy and snowy scenarios can be attributed to the increased exposure to diverse domains, allowing the model to acquire more generalized knowledge and enhance its performance. 
However, the nighttime scenario, which significantly differs from daytime scenarios, faces challenges in extracting relevant knowledge from other scenarios. Nonetheless, even with an increasing number of domains, our method consistently maintains its performance without any decline.

\section{Implementation Details}
\label{sec:details}
In this study, we follow the implementation details from the previous work, P{\O}DA\xspace ~\cite{fahes2023poda}.
Specifically, we utilize the DeepLabv3+ framework~\cite{chen2018encoder} incorporating a backbone model of pre-trained CLIP-ResNet-50\footnote{\url{https://github.com/openai/CLIP}}. For the source domain, the model is trained for 200,000 iterations using randomly cropped 768x768 images. Training is performed with a polynomial learning rate schedule, starting at $lr = 10^-1$ for the classifier, and employing Stochastic Gradient Descent~\cite{bottou2010large} with a momentum of 0.9 and weight decay of $10^-4$. Standard color jittering and horizontal flip augmentations are applied to these crops.

During Stage-1, where the PIN is trained to simulate the target domain feature, we make use of the source feature maps after the first layer. The style parameters $\boldsymbol{\mu}$ and $\boldsymbol{\sigma}$ are represented as 256-dimensional real vectors. The CLIP embeddings are $1024$D vectors. To encode the target descriptions, we adapt the ImageNet templates from~\cite{radford2021learning} and use them in the encoding process of the $\mathsf{TrgPrompt}$.

In the Fine-tuning stage (Stage 2), we start with the pre-trained model on the source domain and focus on fine-tuning the classifier head. This process involves working with augmented PIN features, denoted as $\overline{\mathbf{f}}_{\mathrm{s} \rightarrow \mathrm{t}}$, and continuing the process for 2,000 iterations. 

To evaluate the adaptation performance, we mainly utilize the mean Intersection over Union (mIoU\%) metric. This metric allows us to assess the performance of the models on target images at their original resolutions.

\section{Additional Discussions}
\label{sec:discussion}
\textbf{What's the difference between our and previous settings?}
\begin{table*}[hbt!]
\centering
\scalebox{0.88}{
\tabcolsep12pt

\begin{tabular}{l|c|c|c}
\hline
Setting  & Target Data & Domain-ID  & Model Number          \\ \hline
Standard Unsupervised Domain Adaptation   & Image  &Require   & $N$              \\
One-Shot Unsupervised Domain Adaptation ~\cite{luo2020adversarial}   & Image  &Require       & $N$ \\
Few-Shot Unsupervised Domain Adaptation ~\cite{kalluri2022cluster}   & Image &Require        & $N$ \\
Prompt-driven Zero-shot Domain Adaptation~\cite{fahes2023poda}                           & Scenario Description     &Require   & $N$      \\ \hline
\multicolumn{1}{l|}{Unified Language-driven Zero-shot Domain Adaptation} & Scenario Description    &No Require    & $1$       \\ \hline
\end{tabular}
}
\caption{The difference between our proposed Unified Language-driven Zero-shot Domain Adaptation and related adaptation settings. Target Data means the form of the target domain data. Domain-ID indicates whether the model requires the domain ID during testing. Model Number means for $N$ target domains, the required segmentation head's Number. }\label{tab:settings}
\end{table*}
To facilitate a more comprehensive comparison between our proposed setting and the previous settings, we further analyze the target data format and the number of models required for N target domains. 
As shown in Table~\ref{tab:settings}, Standard Unsupervised Domain Adaptation~(UDA), One-Shot Unsupervised Domain Adaptation~(OSUDA), Few-Shot Unsupervised Domain Adaptation~(FSUDA) all require access to the target domain image. In contrast, Prompt-driven Zero-shot Domain Adaptation~(P{\O}DA\xspace) and our proposed Unified Language-driven Zero-shot Domain Adaptation~(ULDA) only need to access the target domain language description to extract the target domain knowledge, which is practical and less costly.
Besides, compared to the previous settings, only ULDA requires a single model to adapt to diverse target domains without domain-IDs, instead of using domain-specific heads as in previous methods.

\mypara{Why not incorporate TDR to Stage-1?} We present a detailed derivation regarding the question, "Why not incorporate TDR into Stage-1?" as discussed in Section 6 of the main paper here.

For the original source domain feature, we can obtain the corresponding target domain feature $\mathbf{f}_{\mathrm{s} \rightarrow \mathrm{t}}$ through the following formula~(Eq. (1) in the main paper):
\begin{equation*}
\small
\mathbf{f}_{\mathrm{s} \rightarrow \mathrm{t}}=\operatorname{PIN}\left(\mathbf{f}_{\mathrm{s}}, \boldsymbol{\mu}, \boldsymbol{\sigma}\right)=\boldsymbol{\sigma}\left(\frac{\mathbf{f}_{\mathrm{s}}-\mu\left(\mathbf{f}_{\mathrm{s}}\right)}{\sigma\left(\mathbf{f}_{\mathrm{s}}\right)}\right)+\boldsymbol{\mu}.
\end{equation*}
This normalization process transfers features from the source domain to the distribution of the target domain.

For $\mathbf{f}_{\mathrm{s} \rightarrow \mathrm{t}}$, 
we know 
$\displaystyle\frac{\mathbf{f}_{\mathrm{s}}-\mu\left(\mathbf{f}_{\mathrm{s}}\right)}{\sigma\left(\mathbf{f}_{\mathrm{s}}\right)}
$ theoretically follows the distribution $ \mathcal{N}(0,1)$,  thus $\mathbb{E}[\displaystyle\frac{\mathbf{f}_{\mathrm{s}}-\mu\left(\mathbf{f}_{\mathrm{s}}\right)}{\sigma\left(\mathbf{f}_{\mathrm{s}}\right)}
] = 0$, $\mathrm{Var}(\displaystyle\frac{\mathbf{f}_{\mathrm{s}}-\mu\left(\mathbf{f}_{\mathrm{s}}\right)}{\sigma\left(\mathbf{f}_{\mathrm{s}}\right)}
) = 1.$
therefore, we could calculate the $\mathbb{E}(\mathbf{f}_{\mathrm{s} \rightarrow \mathrm{t}})$ and $Var(\mathbf{f}_{\mathrm{s} \rightarrow \mathrm{t}})$ as:

\begin{equation}
\begin{aligned}
\mathbb{E}\left(\mathbf{f}_{\mathrm{s} \rightarrow \mathrm{t}}\right) & =\mathbb{E}\left[\boldsymbol{\sigma}\left(\frac{\mathbf{f}_{\mathrm{s}}-{\mu}\left(\mathbf{f}_{\mathrm{s}}\right)}{\sigma\left(\mathbf{f}_{\mathrm{s}}\right)}\right)+\boldsymbol{\mu}\right] \\
& =\mathbb{E}(\boldsymbol{\mu})+\boldsymbol{\sigma}\left[\left(\frac{\mathbf{f}_{\mathrm{s}}-{\mu}\left(\mathbf{f}_{\mathrm{s}}\right)}{{\sigma}\left(\mathbf{f}_{\mathrm{s}}\right)}\right)\right] \\
& =\boldsymbol{\mu}+\boldsymbol{\sigma} \cdot 0 \\
& =\boldsymbol{\mu} .
\end{aligned}
\end{equation}

\begin{equation}
\begin{aligned}
\operatorname{Var}\left(\mathbf{f}_{\mathrm{s} \rightarrow \mathrm{t}}\right) & =\operatorname{Var}\left[\boldsymbol{\sigma}\left(\frac{\mathbf{f}_{\mathrm{s}}-\mu\left(\mathbf{f}_{\mathrm{s}}\right)}{\sigma\left(\mathbf{f}_{\mathrm{s}}\right)}\right)+\boldsymbol{\mu}\right] \\
& =\operatorname{Var}\left[\boldsymbol{\sigma}\left(\frac{\mathbf{f}_{\mathrm{s}}-\mu\left(\mathbf{f}_{\mathrm{s}}\right)}{\sigma\left(\mathbf{f}_{\mathrm{s}}\right)}\right)\right] \\
& =\boldsymbol{\sigma}^2 \operatorname{Var}\left[\frac{\mathbf{f}_{\mathrm{s}}-\mu\left(\mathbf{f}_{\mathrm{s}}\right)}{\sigma\left(\mathbf{f}_{\mathrm{s}}\right)}\right] \\
& =\boldsymbol{\sigma}^2
\end{aligned}
\end{equation}
Hence, for the simulated $\mathbf{f}_{\mathrm{s} \rightarrow \mathrm{t}}$, we have $std(\mathbf{f}_{\mathrm{s} \rightarrow \mathrm{t}})= \boldsymbol{\sigma}$, $mean(\mathbf{f}_{\mathrm{s} \rightarrow \mathrm{t}})= \boldsymbol{\mu}$. Substituting them into main paper's Eq.~(10) yields:
\begin{equation}
\begin{aligned}
 \mathbf{\widetilde{f}}_{\mathrm{s} \rightarrow \mathrm{t}}&=\beta \left( \boldsymbol{\widetilde{\sigma}} \left( \frac{\mathbf{f}_{\mathrm{s} \rightarrow \mathrm{t}} - \mu\left(\mathbf{f}_{\mathrm{s} \rightarrow \mathrm{t}}\right)}{\sigma\left(\mathbf{f}_{\mathrm{s} \rightarrow \mathrm{t}}\right)} \right) + \boldsymbol{\widetilde{\mu}} \right) + \mathbf{f}_{\mathrm{s} \rightarrow \mathrm{t}} \\
& =\beta \left( \boldsymbol{\widetilde{\sigma}} \left( \frac{\mathbf{f}_{\mathrm{s} \rightarrow \mathrm{t}} - \boldsymbol{\mu}}{\boldsymbol{\sigma}} \right)+\boldsymbol{\widetilde{\mu}}\right)+\boldsymbol{\mu}+\boldsymbol{\sigma}\left(\frac{\mathbf{f}_{\mathrm{s}}-\mu\left(\mathbf{f}_{\mathrm{s}}\right)}{\sigma\left(\mathbf{f}_{\mathrm{s}}\right)}\right) . \\
& =\beta\left(\boldsymbol{\widetilde{\sigma}} \left( \frac{\mathbf{f}_{\mathrm{s}}-\mu\left(\mathbf{f}_s\right)}{\sigma\left(\mathbf{f}_{\mathrm{s}}\right)} \right) +\boldsymbol{\widetilde{\mu}}\right)+\boldsymbol{\mu}+\boldsymbol{\sigma}\left(\frac{\mathbf{f}_{\mathrm{s}}-\mu\left(\mathbf{f}_{\mathrm{s}}\right)}{\sigma\left(\mathbf{f}_{\mathrm{s}}\right)}\right) . \\
& =\left( \frac{\mathbf{f}_{\mathrm{s}}-\mu\left(\mathbf{f}_{\mathrm{s}}\right)}{\sigma\left(\mathbf{f}_{\mathrm{s}}\right)} \right) (\beta \boldsymbol{\widetilde{\sigma}}+\boldsymbol{\sigma})+(\beta\boldsymbol{\widetilde{\mu}}+\boldsymbol{\mu}) . \\
\end{aligned}
\end{equation}

As discussed in the main paper Sec. 6, $\boldsymbol{\widetilde{\sigma}}$ and $\boldsymbol{\widetilde{\mu}}$ are derived by passing text embeddings through a linear layer. The parameters $\boldsymbol{\sigma}$ and $\boldsymbol{\mu}$ are learnable and are designed to simulate features of the target domain. During Stage-1, it is necessary to optimize $\boldsymbol{\mu}$ and $\boldsymbol{\sigma}$ to transform the source domain features into those of the target domain, ensuring alignment with the text embeddings. However, as the text embeddings are directly input into the linear layer to obtain $\boldsymbol{\widetilde{\mu}}$ and $\boldsymbol{\widetilde{\sigma}}$, this process results in $\boldsymbol{\mu}$ and $\boldsymbol{\sigma}$ not being optimized, leading to a trivial solution. Therefore, we may not integrate rectification into Stage-1.
\begin{figure*}
    \centering

    \includegraphics[width=1.0\linewidth]{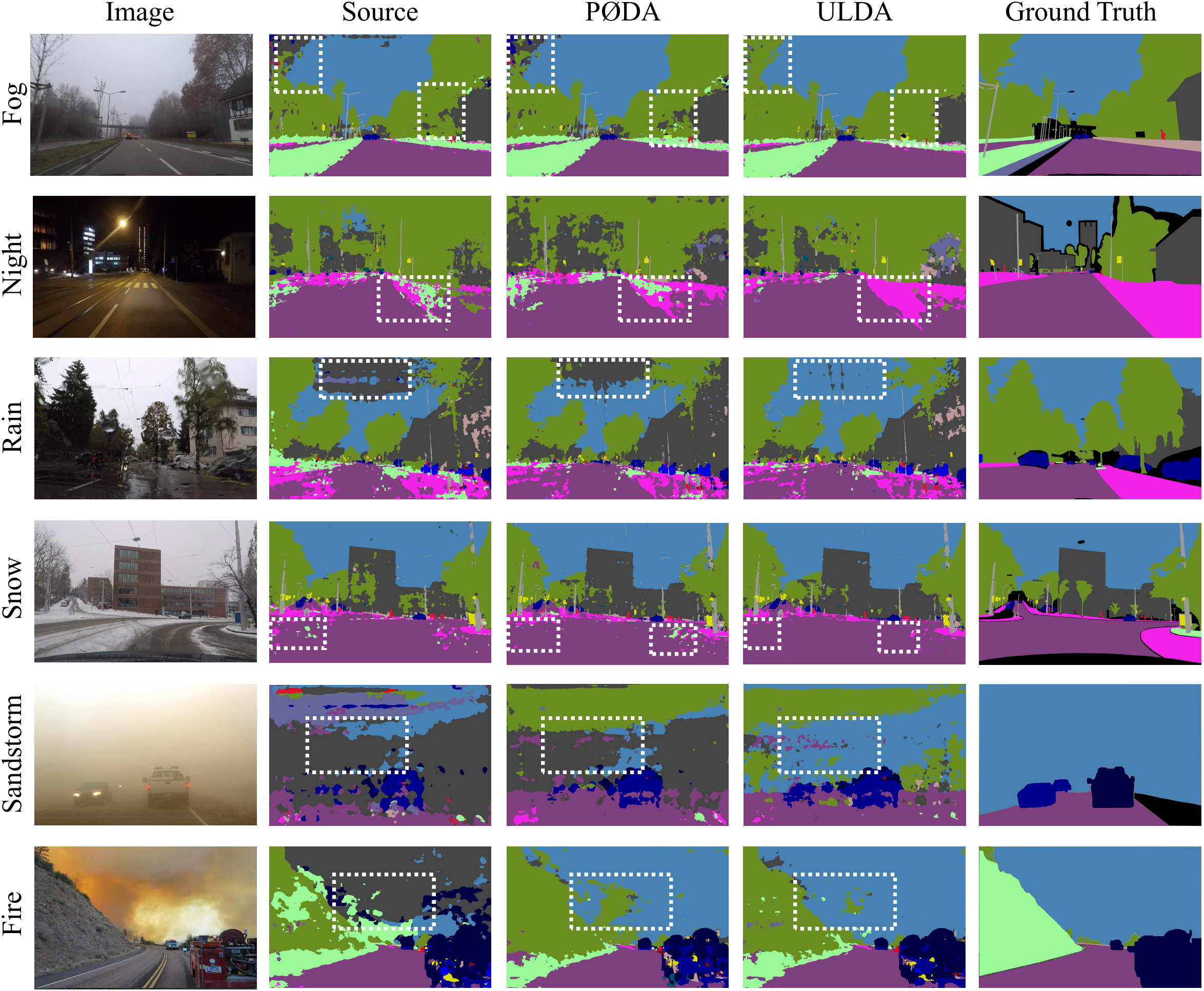}
    \caption{ \textbf{Qualitative Analysis.} We compare the qualitative results of Autonomous Driving scenarios. Only our method employs a single segmentation head across all scenarios, other methods train a segmentation head specifically for each domain.    
    }
    \label{fig:supp_qualitative}
\end{figure*}
\begin{figure*}[ht]
    \centering
    \begin{subfigure}[b]{0.45\textwidth}
        \centering
        \includegraphics[width=0.985\textwidth]{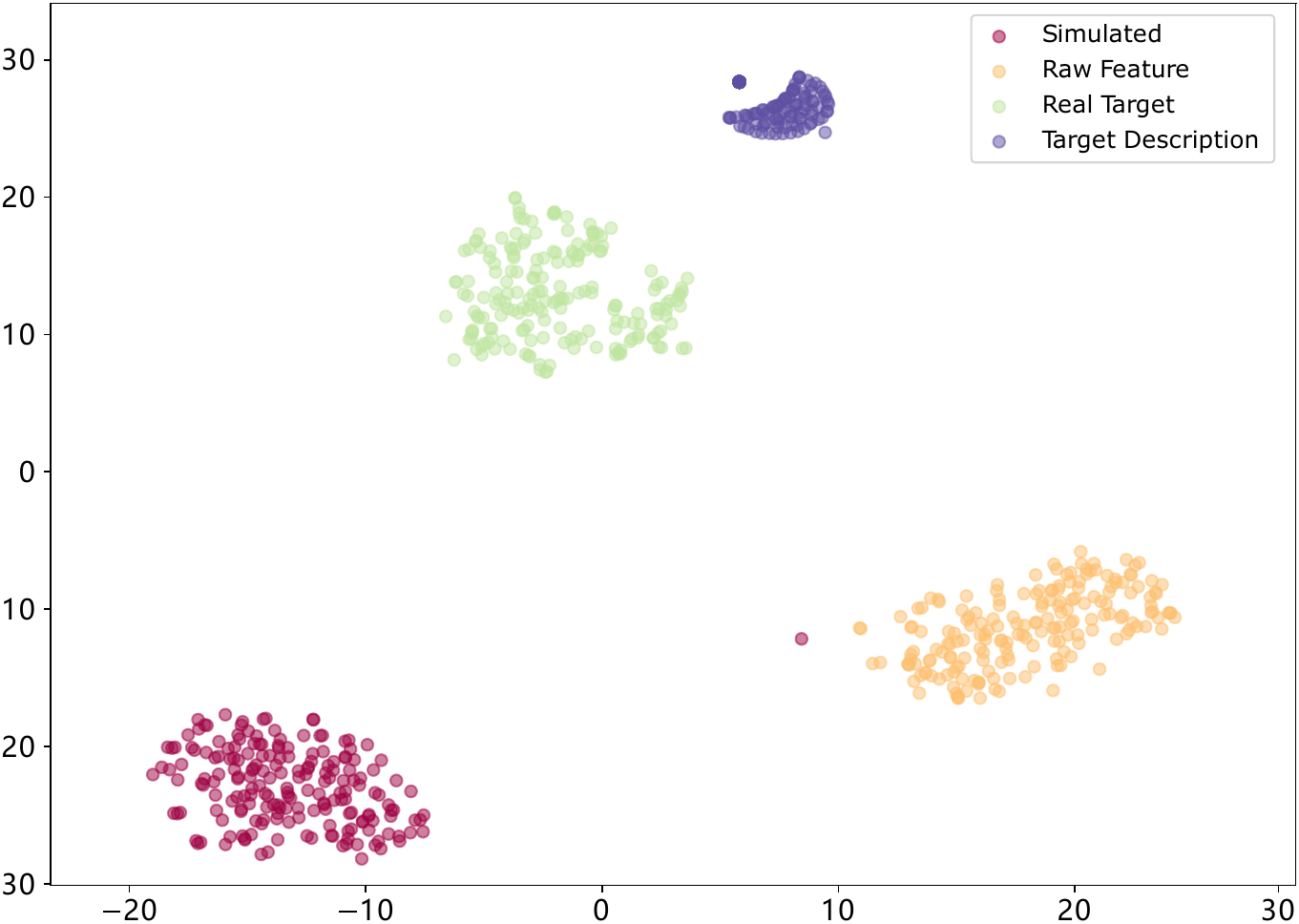}
        \label{fig:real_rec1}
    \end{subfigure}
    \hfill
    \begin{subfigure}[b]{0.45\textwidth}
        \centering
        \includegraphics[width=\textwidth]{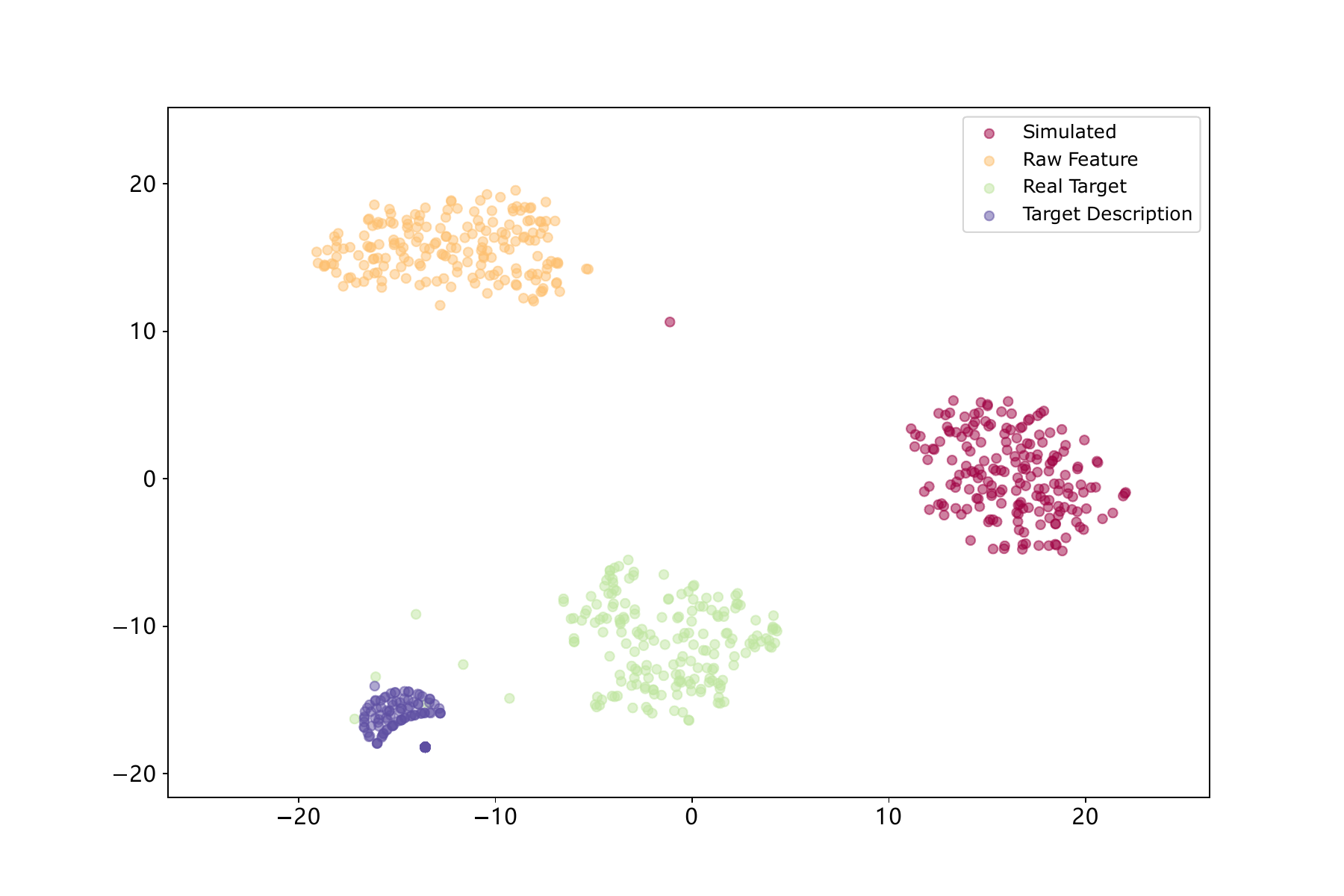}
        \label{fig:real_rec2}
    \end{subfigure}
    \newline
    \begin{subfigure}[b]{0.45\textwidth}
        \centering
        \includegraphics[width=\textwidth]{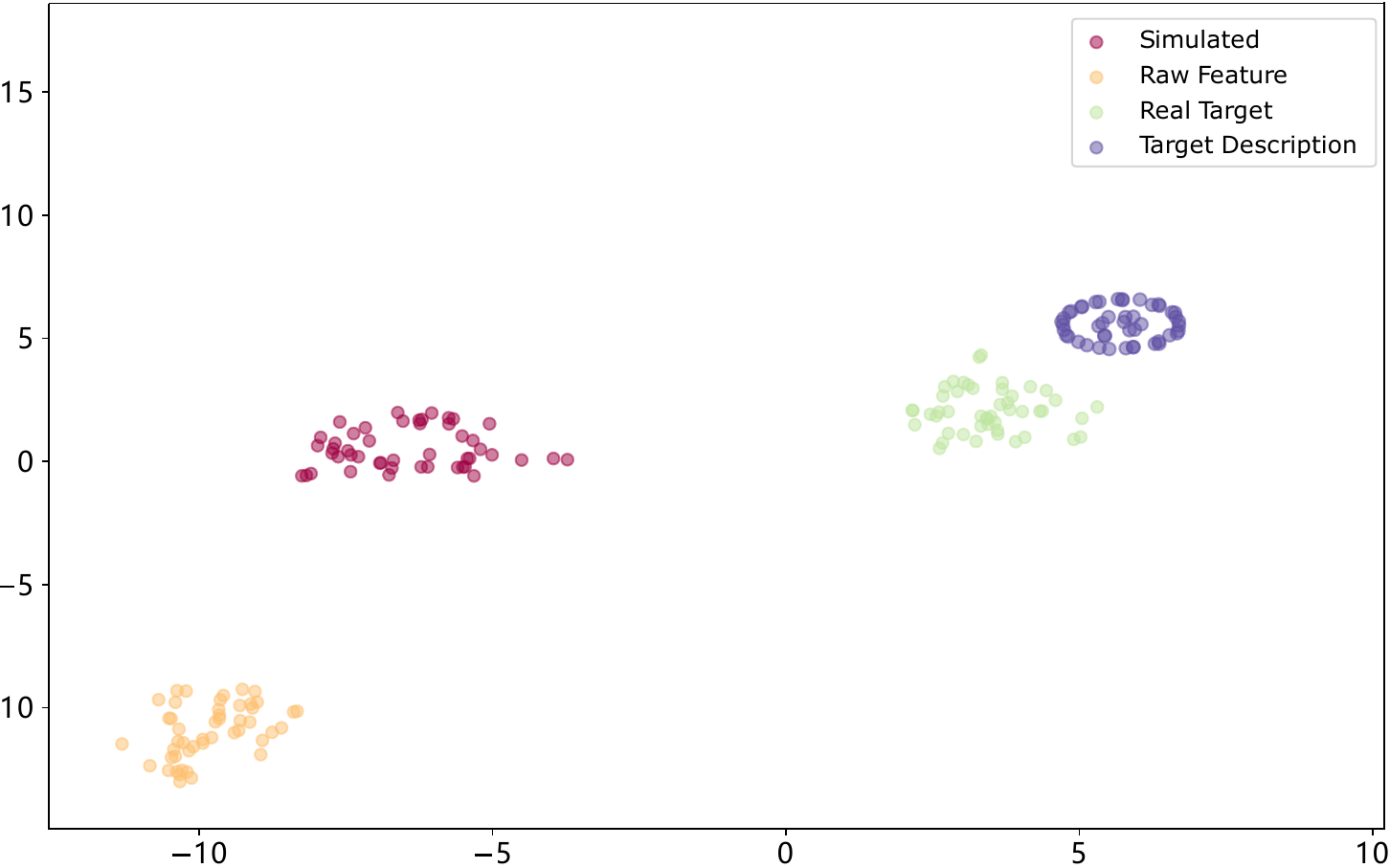}
        \label{fig:real_rec3}
    \end{subfigure}
    \hfill
    \begin{subfigure}[b]{0.45\textwidth}
        \centering
        \includegraphics[width=0.99\textwidth]{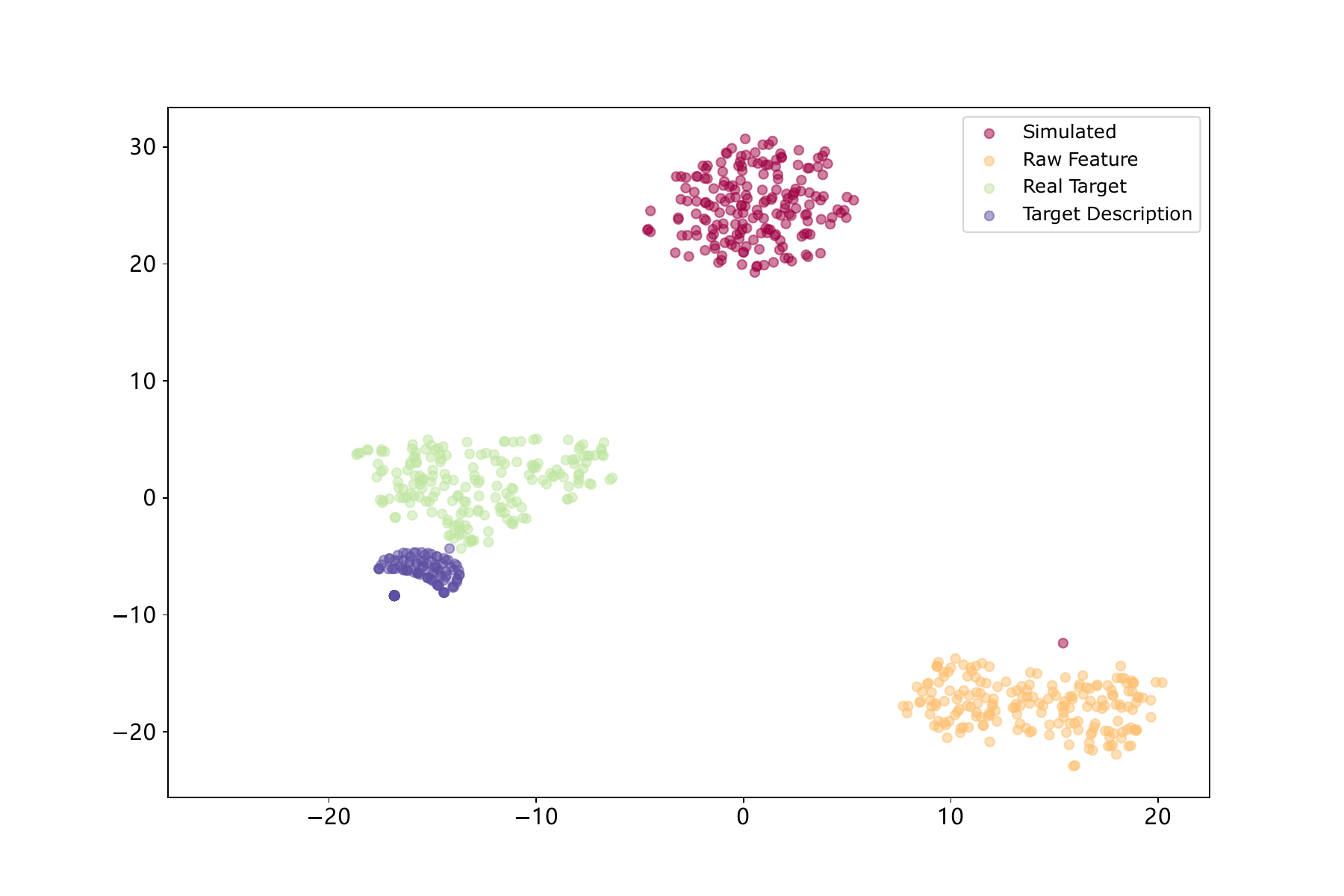}
        \label{fig:real_rec4}
    \end{subfigure}
    \caption{Real data of the Text-Driven Rectifier}
    \label{fig: Real rectify data}
\end{figure*}
\section{Related Work}
\label{sec:related}
\textbf{Unsupervised Domain Adaptation~(UDA)}
In UDA, a model trained on a labeled source domain is adapted to an unlabeled target domain. The majority of the approaches rely on discrepancy minimization~\cite{long2015learning,long2017deep}, adversarial training \cite{ganin2016domain, tsai2018learning} and self-training \cite{zou2019confidence, li2019bidirectional}. These techniques primarily focus on reducing the domain gap at different levels: input \cite{hoffman2018cycada, yang2020fda}, features \cite{ sun2016deep, wang2017deep, zang2023boosting, liu2022unsupervised}, or output \cite{tsai2018learning, vu2019advent}.
However, in real-world scenarios, obtaining a substantial number of target domain images can be challenging, leading to the development of various domain adaptation settings~\cite{liu2023vida, yang2023exploring, liu2023adaptive, ni2023distribution, wang2020tent, wang2022continual, fahes2023poda, panagiotakopoulos2022online}.

Recently, one challenging setting of One-Shot Unsupervised Domain Adaptation (OSUDA) has been proposed. This setting requires models to adapt to a target domain with access to only one image from that domain. To the best of our knowledge, only three studies focusing on semantic segmentation within this context have been documented \cite{luo2020adversarial, wu2022style, benigmim2023one}.
Luo et al. \cite{luo2020adversarial} highlight the limitations of traditional UDA methods when limited to a single unlabeled target image. They propose a style mining algorithm that combines a stylized image generator with a task-specific module to prevent overfitting. 
In contrast, Wu et al. \cite{wu2022style} introduce a novel approach named style mixing and patch-wise prototypical matching (SM-PPM). This method involves blending the features of a source image with those of the target linearly, and employing patch-wise prototypical matching to mitigate negative adaptation \cite{li2020content}.
Benigmim et al.~\cite{benigmim2023one} advance the field by introducing Stable Diffusion and DreamBooth techniques. They extract domain knowledge from the pre-trained model, enabling the transfer of source images to the target image.

For the more challenging setting Zero-Shot Unsupervised Domain Adaptation, where no target image is available, Lengyel et al. \cite{lengyel2021zero} explore day-to-night domain adaptation. They introduce the Color Invariant Convolution Layer (CIConv) to achieve network invariance under varying lighting conditions. However, this approach heavily relies on the physics prior knowledge and is specifically tailored to a certain type of domain gap.

More recently, Fahes et al.~\cite{fahes2023poda} introduced a new setting named Prompt-driven Zero-shot Domain Adaptation. This setting, which does not allow access to the target domain data, leverages natural language descriptions of the target domain to adapt the model to new environments. P{\O}DA\xspace utilizes CLIP to extract the target domain knowledge embedded within these natural descriptions. It employs a two-stage process to simulate target domain features, effectively addressing the domain gap. 

\mypara{Text-driven Vision Models.}
Recent advancements in contrastive image-language pretraining have led to significant achievements in multimodal learning across various tasks such as zero-shot classification~\cite{radford2021learning}, multi-modal retrieval~\cite{jia2021scaling}, visual question answering~\cite{li2021align}, and have facilitated extensive work on Multimodal Large Language Models~\cite{li2023blip,regionblip, liu2024visual,lai2023lisa,li2023llama,li2023manipllm,yang2023lidar}. These developments have paved the way for modifying images using textual descriptions, bridging the previously challenging gap between visual and linguistic representations. 
For text-guided style transfer, CLIPstyler~\cite{kwon2022clipstyler} diverges from relying on a generative process. This methodology offers a more realistic approach as it is not restricted to a specific training distribution, yet it simultaneously poses challenges due to the necessity of utilizing the encapsulated information within the CLIP latent space. The absence of a direct mapping between image and text representations necessitates regularization to effectively extract useful information from text embeddings. In this context, CLIPstyler optimizes a U-net autoencoder to preserve content, while varying the output image embedding in the CLIP latent space during the optimization process.

\mypara{Semantic Segmentation} which involves the classification of each pixel in an image, is a crucial task in computer vision. Several notable contributions in this domain have been introduced~\cite{medical2022,chen2017deeplab, zhao2017pyramid, xie2021segformer,prune2021jianhui,ftn,10130611,sdc,fsformer,cac_aaai,decouplenet,apd,wang2023handwriting}.
Although these methods achieve impressive results, they often require considerable amounts of pixel-level annotated data, which can be a laborious and time-consuming task to collect and annotate. Additionally, they may face difficulties in effectively generalizing when deployed in new domains.

To address these challenges, numerous few-shot~\cite{pfenet, gfsseg,peng2023hierarchical, peng2024oa, pfenet++} and semi-supervised~\cite{jiang_semi, cac_cvpr, zhang2024boundary,zhang2024seine} methods have been proposed.
Recent research has primarily focused on addressing these challenges by employing domain adaptation strategies. For instance, in~\cite{yang2020fda}, a method is proposed that swaps the low-frequency spectrum to align the source and target domains. Another approach~\cite{tranheden2021dacs} involves mixing images from both domains along with their corresponding labels and pseudo-labels. Besides,~\cite{wu2021dannet} utilizes adversarial learning to train a domain adaptation network specifically for nighttime semantic segmentation. Furthermore,~\cite{hoyer2022daformer} introduces a novel model and training strategies to enhance training stability and mitigate overfitting to the source domain. Lastly,~\cite{hoyer2023mic} employs masking of the target images to enable the model to learn spatial context relations of the target domain, providing additional clues for robust visual recognition.
However, these methods both require access to the image of the target domain, which may not be feasible in some reality scenarios. Thus, we propose the Unified Language-driven Zero-shot Domain Adaptation to address this problem.
\section{Qualitative Analysis}
\label{sec:qualitative}
\mypara{Qualitative Analysis on Autonomous Driving scenarios} As shown in Fig.~\ref{fig:supp_qualitative}, we compare our proposed ULDA with the previous method on Autonomous Driving scenarios. Specifically, our visualization results are generated using a comprehensive all-in-one head, while the visualization results of the previous state-of-the-art (SOTA) method are generated using specific heads in each domain.
The figures demonstrate that our method performs well on various objects, such as sidewalks and sky cars. This further reinforces the effectiveness of our approach.

\mypara{Real data of the Text-Driven Rectifier}
In Section 4.3 of the main paper, we assert that using simulated target domain features to fine-tune the segmentation head may result in persistent discrepancies between the simulated features and the actual target domain features. To support this claim, we present real data in Figure~\ref{fig: Real rectify data}.
The relationships among the simulated feature, raw feature, real target feature, and the target description cluster demonstrate that due to the limitations of the simulating process, access to target domain data is restricted, resulting in the failure to capture the simulated features accurately. These findings highlight the significance of incorporating the Text-Driven Rectifier into the fine-tuning process. By doing so, it encourages closer alignment between the simulated features and the target features.

\end{document}